\RequirePackage{amsthm}

\documentclass[sn-nature,iicol]{sn-jnl}

\usepackage{graphicx}%
\usepackage{multirow}%
\usepackage{amsmath,amssymb,amsfonts}%
\usepackage{amsthm}%
\usepackage{mathrsfs}%
\usepackage[title]{appendix}%
\usepackage{xcolor}%
\usepackage{textcomp}%
\usepackage{manyfoot}%
\usepackage{booktabs}%
\usepackage{algorithm}%
\usepackage{algorithmicx}%
\usepackage{algpseudocode}%
\usepackage{listings}%
\usepackage[firstpageonly=true]{draftwatermark}%

\theoremstyle{thmstyleone}%
\theoremstyle{thmstyletwo}%

\theoremstyle{thmstylethree}%

\begin{document}
\SetWatermarkAngle{0}
\SetWatermarkColor{black}
\SetWatermarkLightness{0.5}
\SetWatermarkFontSize{9pt}
\SetWatermarkVerCenter{50pt}
\SetWatermarkText{\parbox{40cm}{%
\centering This version of the article has been accepted for publication, after peer review, but it is not\\
\centering the Version of Record and does not reflect post-acceptance improvements, or any
corrections. \\
\centering The Version of Record is available online at: \\
\centering Patni, S.P., Stoudek, P., Chlup, H. and Hoffmann, M., 2024.\\
\centering Online elasticity estimation and material sorting using standard robot grippers. \\
\centering  The International Journal of Advanced Manufacturing Technology, pp.1-19. (C) Springer \\
\centering \url{https://doi.org/10.1007/s00170-024-13678-6}
}}

\title[Stiffness Estimation using Robotic Grippers]{Online Elasticity Estimation and Material Sorting Using Standard Robot Grippers}


\author*[1]{\fnm{Shubhan P.} \sur{Patni}}\email{patnishu@fel.cvut.cz}
\equalcont{These authors contributed equally to this work.}

\author[1]{\fnm{Pavel} \sur{Stoudek}}
\equalcont{These authors contributed equally to this work.}

\author[2]{\fnm{Hynek} \sur{Chlup}}

\author[1]{\fnm{Matej} \sur{Hoffmann}}

\affil*[1]{\orgdiv{Department of Cybernetics, Faculty of Electrical Engineering}, \orgname{Czech Technical University in Prague}}

\affil[2]{\orgdiv{Department of Mechanics, Biomechanics and Mechatronics, Faculty of Mechanical Engineering}, \orgname{Czech Technical University in Prague}}


\abstract{Stiffness or elasticity estimation of everyday objects using robot grippers is highly desired for object recognition or classification in application areas like food handling and single-stream object sorting. However, standard robot grippers are not designed for material recognition. 
We experimentally evaluated the accuracy with which material properties can be estimated through object compression by two standard parallel jaw grippers and a force/torque sensor mounted at the robot wrist, with a professional biaxial compression device used as reference. Gripper effort versus position curves were obtained and transformed into stress/strain curves. The modulus of elasticity was estimated at different strain points and the effect of multiple compression cycles (precycling), compression speed, and the gripper surface area on estimation was studied. Viscoelasticity was estimated using the energy absorbed in a compression/decompression cycle, the Kelvin-Voigt, and Hunt-Crossley models. We found that: (1) slower compression speeds improved elasticity estimation, while precycling or surface area did not; (2) the robot grippers, even after calibration, were found to have a limited capability of delivering accurate estimates of absolute values of Young’s modulus and viscoelasticity; (3) relative ordering of material characteristics was largely consistent across different grippers; (4) despite the nonlinear characteristics of deformable objects, fitting linear stress/strain approximations led to more stable results than local estimates of Young's modulus; (5) the Hunt-Crossley model worked best to estimate viscoelasticity, from a single object compression. 
A two-dimensional space formed by elasticity and viscoelasticity estimates obtained from a single grasp is advantageous for the discrimination of the object material properties. We demonstrated the applicability of our findings in a mock single stream recycling scenario, where plastic, paper, and metal objects were correctly separated from a single grasp, even when compressed at different locations on the object. The data and code are publicly available.}

\maketitle
\section{Introduction}
\label{sec:intro}

Although visual object recognition has seen tremendous progress in recent years, not all object properties can be perceived using distal sensing. In particular, physical object properties like stiffness, roughness, or mass are better perceived through manipulation. Object recognition from  haptic exploration can be more robust as it is insensitive to lighting conditions and object attributes that may be irrelevant (e.g., color). 

Sanchez et al.~\cite{sanchez2018robotic} provide a survey of robotic manipulation and sensing of deformable objects. Objects are considered deformable if they have no compression strength (ropes and clothes) or have a large strain\footnote{For linear elasticity, this implies small Young's modulus, e.g. less than 10000 kPa.}. Additionally, classification based on geometry is presented. In this work, we leave objects of Type I-III (linear, planar, cloth-like) aside and focus on Type IV: triparametric objects---solid objects such as sponges or plush toys, which are also the least researched object type \cite{sanchez2018robotic}. Here we extend the elasticity range and test much less elastic but still deformable materials like cardboard, plastic, and metal. 

In this work, we specifically focus on material elasticity and viscoelasticity, and to what extent it can be estimated online using common robot equipment like 2-finger grippers. Grippers are designed for grasping rather than stiffness estimation, and often lack tactile sensors at the fingertips, hence their ability to provide accurate estimates of physical quantities like the modulus of elasticity is limited. With the help of a professional biaxial compression testing device, we systematically assess the performance of these grippers and a force sensor at the robot wrist, under different settings. 

The primary application areas are in fruit ripeness estimation and single stream recycling. Research on automatic fruit ripeness estimation (tomatoes, nectarines \cite{lin2023non}, apples, strawberries \cite{ribeiro2020fruit}) or delicate edible manipulation \cite{cardin2023gripper} featured special-purpose soft or variable stiffness grippers and tactile sensors. Automatic waste separation often relies on distal sensing (\cite{lubongo2022assessment} discuss Near InfraRed (NIR), X-ray Fluorescence (XRF), and vision sensing for plastic separation); Chin et al.~\cite{chin2019automated} presented a special-purpose gripper prototype for haptic discrimination of plastic, paper, and metal.
Additional application areas where deformable objects need to be handled include object picking for manufacture~\cite{papadopoulos2023deformable, zhang2023determination}, virtual/augmented reality \cite{liu2018variable}, and medical diagnosis through palpation of subcutaneous tissue \cite{scimeca2022action}. 
Our work differs in that we evaluate the potential of the use of standard industrial robot grippers for online elasticity estimation and material recognition from a single grasp.

This article is structured as follows. After reviewing related work, the essentials about the physics of deformation are explained (Section~\ref{sec:physics_of_deformation}). Sample objects and the measuring devices are described in the Experimental Setup section, followed by Experiments and Results. We close with Discussion, Conclusion, and Future Work.

The dataset including raw as well as processed data collected during the experiments in this work is publicly available \cite{osf,dataset}. Two accompanying videos are available: Part 1, on the results of using different models to estimate elasticity and viscoelasticity \url{https://youtu.be/8DMMdEezC1M}; Part 2, showing the application of stiffness estimation to sort materials in a single-stream recycling scenario: \url{ https://youtu.be/XHiNKFZ158o}.

\section{Related work}
\label{sec:rel_work}
Recent surveys of haptic or tactile robot perception are provided by \cite{li2020review,luo2017robotic}. Li et al.~\cite{li2020review} list the object and material properties that can be extracted via tactile perception: stiffness, friction, surface texture, thermal conductivity, and adhesion. However, Luo et al.~\cite{luo2017robotic} point out a gap between rapid development of tactile sensing hardware and insufficient knowledge about how to interpret the sensory information obtained.
In this paper, we try to reduce this gap by exploring the data streamed from the sensors and extracting material properties that may improve the interactions between robots and objects. We focus on stiffness/elasticity and how it can be perceived through a pinch grasp (squeezing). For deformable objects, the same action also allows us to perceive another material property: viscoelasticity.  
The next subsections provide an overview of the latest research in estimating stiffness and viscoelasticity in a robotic setup, i.e. without the use of industrial precision measurement instruments but rather with sensors commonly integrated with robots---RGB-D cameras and tactile sensors that measure reaction force.

\subsection{Object Discrimination from Tactile Feedback}
 Spiers et al.~\cite{spiers2016single} used a two-finger compliant gripper with force sensors to classify objects of various shapes, sizes, and stiffness. Random forests were used for classification and the importance of different extracted features was evaluated. Delgado et al.~\cite{delgado2015tactile} used the Shadow Hand to grasp deformable objects and computed the deformability ratio to infer the maximum force allowed to be exerted on the object. Hosoda and Iwase~\cite{hosoda2010robust} used a human-like robot hand with artificial muscles and soft skin with tactile receptors as input to a recurrent neural network with context nodes for object classification. Industrial research in the area is focused towards recognizing objects for improved grasping and manipulation.
 Gemici and Saxena~\cite{gemici2014learning} learned haptic representations of food items. Four actions were used to explore 6 object characteristics (hardness, plasticity, elasticity, tensile strength, adhesiveness, brittleness). Scimeca et al.~\cite{scimeca2019non} used a 2-finger gripper for mango ripeness estimation. Chin et al.~\cite{chin2019automated} implemented a custom-made soft 2-finger gripper with pressure and strain sensors to automated recycling separation (paper, plastic, metal). Finally, most related to this work, Wang et al.~\cite{wang2021tactual} used the deformation cues from pinch grasps of different soft objects to classify them using Functional Principal Component Analysis (FPCA features).

\subsection{Estimating Elasticity}
\label{subsec:lit_review_estimation}

Smardzewski et al.~\cite{smardzewski2008nonlinear} employed a professional axial compression device for estimating the compressive modulus of elasticity, in line with the corresponding standards~\cite{iso3386polymeric}. Such measurement is very different from practical robotics as grippers are much less accurate and the shape of real-world objects complicates measurements if pure material elasticity.

Some works employ the Finite Element Method (FEM) to model the object based on deformation data. Frank et al.~\cite{frank2010learning} estimated the stiffness by capturing the object via RGB-D camera, creating a three dimensional finite element mesh in simulation software, and then numerically optimizing the stiffness parameter of the mesh by minimizing the error between the simulated and observed deformations. 
Visual and haptic information was also combined by Longhini et al.~\cite{longhini2022edo, longhini2023elastic}. Marker nodes placed on the surface of the deforming object were tracked visually to learn deformation models of various textiles for control and manipulation. Stress-strain models were used to gain information about deformable materials without explicitly estimating extracting stiffness/elasticity values.
Narang et al.~\cite{narang2021sim} used feedback from the BioTac tactile sensor as input to a multi-layer perceptron network to predict the deformation of the target object in simulation. The network was trained to convert feedback to deformation and vice versa via a latent space representation. This solution had been previously applied to a medical scenario in \cite{bickel2009capture} for estimating the variable stiffness of soft tissue. 
However, the finite element method comprises optimizing the target parameter for a large number of nodes. It is computationally time consuming and cannot be employed for quick, online estimation. Further, these works press the object against a barrier with a fingertip, while in this work, we employ `squeezing' or pinching between the gripper jaws.

Zaidi et al. \cite{zaidi2017model} employed a fingertip contact model simulating the deformation of the target object. The stiffness and damping were computed at the contact point and used as parameters in the deformation simulation to estimate grasp stability.
Haddadi et al.~\cite{haddadi2012real} used the Hunt-Crossley Dynamic model for real-time identification of contact environments. This method is quicker as compared to finite element methods because it regards the object as a whole rather than as a mesh of a large number of nodes. Only three parameters have to be computed from collected force vs. deformation data. Further, the feedback is only from the gripper and attached force sensors (no visual input).
Bednarek et al.~\cite{bednarek2019robotic} did not use a robotic hand or gripper, but instead use a spherical tip with the OptoForce 3-axis optical force sensor. The time series from this sensor was processed using an LSTM neural network employed for material classification.
Yao et al.~\cite{yao2023estimating} used the compression action to create three-dimensional stiffness maps of household objects.
 
\subsection{Our Contribution}
\begin{enumerate}
    \item We use three different robotic devices (two parallel jaw grippers and one force/torque sensor at the robot flange) and systematically study the effect of the characteristics of the device itself and different parameters of the object compression process on the quality of the mechanical response curves.
    \item We test a large set of deformable objects and employ a professional biaxial compression device to provide reference values.
    \item We assess the potential and limitations of robot grippers to gauge material elasticity---absolute values of Young modulus as well as relative ordering of objects by their elasticity.
    \item We assess the accuracy of different methods for estimating viscoleasticity.
    \item We consolidate our findings into a mock waste sorting scenario where a single grasp by the robot gripper suffices to separate recyclable materials.
    \item The collected dataset including the code for data processing is publicly available at \url{https://osf.io/gec6s/}. 
\end{enumerate}

The unique contribution of this work lies in a systematic experimental assessment of the potential and limitations of using standard robot grippers for elasticity estimation and object discrimination based on material composition. Related research on haptic elasticity estimation or material recognition has relied on custom-built complex grippers with dedicated sensors \cite{lin2023non,ribeiro2020fruit,cardin2023gripper,chin2019automated}. We compare several methods for gauging elasticity and viscoelasticity from the experimentally obtained stress/strain response curves and conclude with practical guidelines on how the material properties can be best estimated from a single compression cycle (grasp), culminating in a demonstrator resembling a waste sorting scenario.

\section{Physics of deformation}
\label{sec:physics_of_deformation}

\subsection{Elasticity}

If a force is exerted on an object, the object deforms and it's dimensions change.\footnote{This section draws on \cite[Ch. 9]{giancoli1995physics}} Depending on the direction of the force, the object either compresses or elongates. In this work, we focus on axial compression as this mode of testing objects/materials applies to robot grippers. If the amount of compression, $\Delta L$, is small compared to the dimension of the object, $\Delta L$ is proportional to the force exerted. This is sometimes referred to as Hooke's law and can be written as 
\begin{equation}
\label{eqn:HookesLaw}
    F = k \Delta L,
\end{equation}
where F represents the force applied on the object, $\Delta L$ is the change in length, and \textit{k} is a proportionality constant. This linear relationship is a good approximation for many common materials, but applies only up to a certain $\Delta L$ called the \textit{proportional limit}. With additional compression, the linear relationship does not hold anymore, but up to the \textit{elastic limit}, if the force is released, the object still fully returns to its original dimensions.
The \textit{strain} is the ratio of the change in length $\Delta L$ to the original length $L_0$. It is proportional to the force applied and the original length, but also inversely proportional to the cross-sectional area. To eliminate the object's size or shape, a constant of proportionality that depends only on the object material is desired. This is the \textit{elastic modulus} or \textit{Young's modulus}, \textit{E}:
\begin{equation}
    E = \frac{stress}{strain} = \frac{F/A}{\Delta L / L_0},
\label{eq:Young}
\end{equation}
where \textit{stress} is the force (\textit{F}) per unit area (\textit{A}) and has thus units $Nm^{-2}$, or Pascals. For the soft materials in this work, we will use kiloPascals (kPa). Strain is a dimensionless quantity (no units).  

\subsection{Viscoelasticity}
\label{subsec:DynMod_Hyst}
Soft objects like the polyurethane foams, for example, have an elastic component and a viscoelastic component. Due to molecular rearrangement, energy as heat is dissipated when a load is applied and then removed, i.e. during a compression and decompression cycle. The strain rate is also time-dependent. These properties can be characterized by the hysteresis in the stress-strain curve or by models taking into account both elasticity and viscosity. These will be detailed below.

\subsubsection{Calculating Viscoelasticity from Loop Enclosed Area}
\label{subsubsec:Hysteresis_Area}
\begin{figure}
    \centering
    \includegraphics[width=0.5\textwidth]{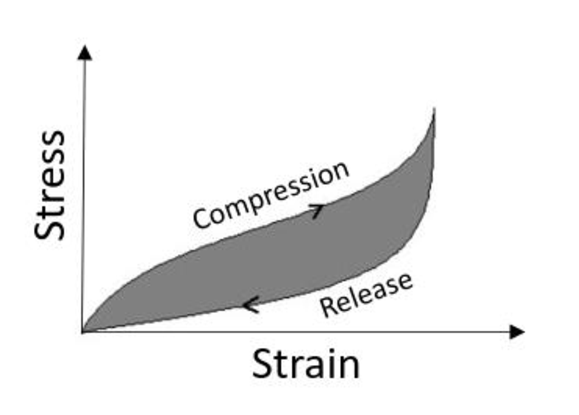}
    \caption{Sample stress-strain curve with hysteresis loop charactersistic of viscoelasticity.}
    \label{fig:DemoHys}
\end{figure}

The ``paths'' for compression and decompression on the stress/strain plot are different for deformable objects. There is an area enclosed by these paths---the hysteresis loop---which corresponds to the amount of energy lost (as heat) as shown in Fig.~\ref{fig:DemoHys}. By plotting the energy lost versus the speed of compression/release we can obtain the viscoelastic coefficient $\eta$.

\subsubsection{Kelvin-Voigt Model}

The Kelvin-Voigt model is a simple model that views every object as a spring and damper system arranged in parallel. The equation used to describe objects is a linear second order differential equation:

\begin{equation}
F(x(t),\dot{x}(t))=Kx(t) + \eta \dot{x}(t)
    \label{eqn:KVModel}
\end{equation}

where $F$ is the applied force, $K$ is the stiffness/elasticity parameter, $\eta$ is the damping/hysteresis parameter, $x$(t) and $\dot{x}$(t) are the deformation and rate of deformation respectively.
Once a cycle of compression and release data has been collected in the form of deformation and force feedback, this data can be fit into Eqn.~(\ref{eqn:KVModel}) using multi-variable linear regression to obtain values of $K$ and $\eta$.

\subsubsection{Hunt-Crossley Model}
The Hunt-Crossley model expands upon the simple Kelvin-Voigt model by adding nonlinear power terms to the stiffness and damping components. Eqn.~\ref{eqn:KVModel} is modified as
\begin{equation}
    F(x(t),\dot{x}(t))=Kx^{n}(t) + \eta x^{n}(t)\dot{x}(t)
    \label{eqn:HCModel}
\end{equation}
\label{subsec:HCModel}
The added power terms $x^n(t)$ accommodate the nonlinearities of the material. The term $n$ is a new parameter obtained as a result of the model. Since it is an exponential, instead of fitting data on this equation, we fit data on the logarithmic form:
\begin{equation}
    \log(F(x,\dot{x}))=\log(K)+n\log(x)+\log[1+\frac{\eta \dot{x}}{K}]
    \label{eqn:HCModel_log}
\end{equation}

\section{Experimental setup}
\label{sec:materials_methods}
The set of object samples is described, followed by the robot grippers and a professional measuring device. Additional details about the experimental procedure and how the robot grippers were commanded can be found in a student thesis \cite{Stoudek2020}. The code used to command the grippers and preprocess data is available at \cite{gitlab-repo-ipalm}.

\subsection{Objects set}
\label{subsec:methods_objects}
We considered four sets of deformable objects. The first set, \textit{cubes and dice}, consisted of 7 cuboid objects with a different size and degree of softness. The second set, \textit{polyurethane foams set}, consisted of 20 polyurethane foam blocks of similar size, along with reference values for elasticity and density provided by the manufacturer. A third, mixed, set was formed from these two.
For these object sets, the deformation can be regarded as elastic---objects returned to their undeformed shapes once the external force is removed. For some objects, this recovery was slow and they can be regarded as viscoelastic.
A fourth set was formed from groceries for the final waste sorting demonstrator. Several of these objects were compressed beyond their elastic limit.

\begin{figure}
    \includegraphics[width=0.48\textwidth]{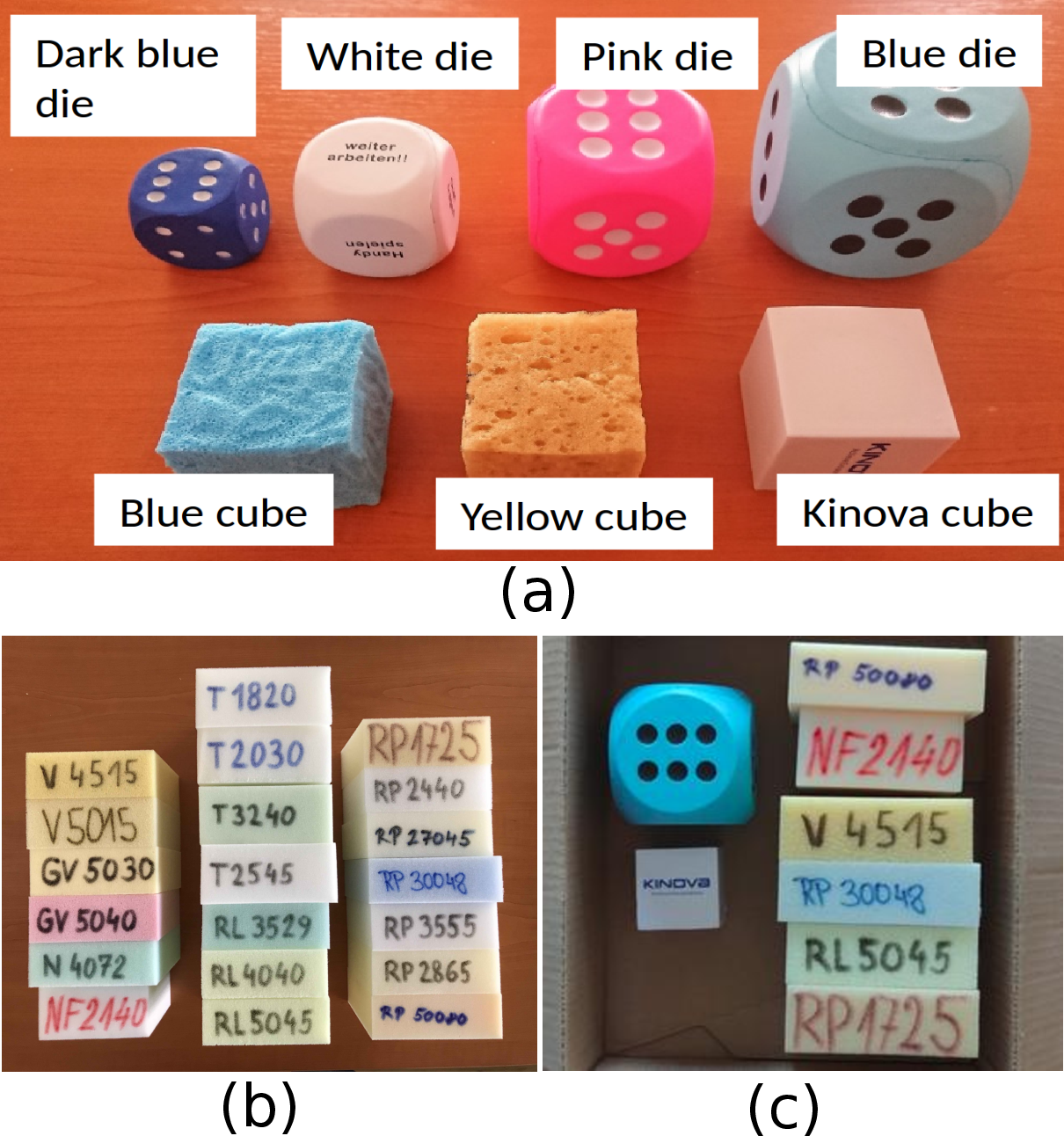}
    \caption{(a) Cubes and Dice set; (b) Polyurethane foams set; (c) Mixed set.}
    \label{fig:object_sets}
\end{figure}

\paragraph{Cubes and dice set}
This set of 7 soft objects is visualized in Fig.~\ref{fig:object_sets}(a) together with their names. Their dimensions are listed in Table~\ref{tab:cubes_dimensions}. The ``blue cube'' is composed of the same material like the ``blue die''---it has been cut out from another exemplar of the same object. ``White die'', ``Kinova cube'', ``Yellow cube'', and ``Blue cube'' have roughly same dimensions but different material composition. The dataset was deliberately designed in this way such that the effects of material elasticity and object shape and size would be apparent in the results.  

\begin{table}[h]
    \caption{Cubes and dice set -- dimensions.}
    \label{tab:cubes_dimensions}
    \begin{tabular}{@{}ll@{}}
        \toprule
        Description            & Dimensions [mm] \\
        \midrule
        Kinova cube            & 56x56x56        \\
        Blue cube              & 56x56x56        \\
        Yellow cube            & 56x56x56        \\
        Blue die               & 90x90x90        \\
        White die              & 59x59x59        \\
        Pink die               & 75x75x75        \\
        Darkblue die           & 43x43x43        \\
        \bottomrule
    \end{tabular}
\end{table}

\paragraph{Polyurethane foams set}
To complement the ``ordinary objects set'', we tested on another set of objects: 20 polyurethane foams (Fig.~\ref{fig:object_sets}(b)). 
These were samples provided by a manufacturer of mattresses and other foam products. The set includes hard insulation foams, memory foams, mattress foams and soundproof foams. Compared to the ordinary soft objects, ``cubes and dice'', this dataset is much more uniform---both in terms of object dimensions and elasticity. The first two letters code for the standardized material type. For example, ``RP'' stands for Richfoam Polyether (STANDARD 100 by OEKO-TEX); the ``T'' series are polyurethane foams used to make furniture (hyper elastic polyurethane furniture foams \cite{smardzewski2008nonlinear}). Additionally, the complete code also contains density and one measure of the material's elasticity: the compression stress value at 40\% compression ($CV_{40}$ \cite{iso3386polymeric}) -- see Table~\ref{tab:foams}. However, we could not directly use these elasticity values as reference because the testing conditions prescribed by the ISO standard \cite{iso3386polymeric} cannot be met with robot grippers; moreover, these values have been obtained from samples tested within 72 hours after manufacture. 

\begin{table}[htb]
    \caption{Polyurethane foams set. Label -- code supplied by manufacturer. Density is expressed in ${kg \cdot m^{-3}}$. Elasticity is the $CV_{40}$ in kPa -- compression stress value at 40\% strain~\cite{iso3386polymeric}.}
    \label{tab:foams}
    \begin{tabular}{@{}llll@{}}
        \toprule
        Label    & Dimensions [mm] & Density & $CV_{40}$ \\
        \midrule
        V4515   & 118x120x40      & 45         & 1.5        \\
        V5015   & 119x120x42      & 50         & 1.5        \\
        GV5030  & 118x119x40      & 50         & 3.0        \\
        GV5040  & 118x118x39      & 50         & 4.0        \\
        N4072   & 118x117x37      & 40         & 7.2        \\
        NF2140  & 105x100x50      & 21         & 4.0        \\
        T1820   & 125x125x50      & 18         & 2.0        \\
        T2030   & 125x120x40      & 20         & 3.0        \\
        T3240   & 123x123x50      & 32         & 4.0        \\
        T2545   & 125x125x50      & 25         & 4.5        \\
        RL3529  & 119x118x40      & 35         & 2.9        \\
        RL4040  & 117x120x40      & 40         & 4.0        \\
        RL5045  & 118x118x39      & 50         & 4.5        \\
        RP1725  & 118x120x41      & 17         & 2.5        \\
        RP2440  & 118x120x38      & 24         & 4.0        \\
        RP27045 & 117x119x39      & 270        & 4.5        \\
        RP30048 & 123x121x39      & 300        & 4.8        \\
        RP3555  & 117x119x39      & 35         & 5.5        \\
        RP2865  & 118x118x38      & 28         & 6.5        \\
        RP50080 & 121x118x39      & 500        & 8.0        \\
        \botrule
    \end{tabular}
\end{table}

\paragraph{Mixed set (Cubes, dice, foams)}
An additional object set was formed from the previous sets, taking two ``cubes and dice'' (``Blue die'' and ``Kinova cube'') and 6 polyurethane foams  -- Fig.~\ref{fig:object_sets}(c). This set of 8 objects was tested on a professional biaxial compression setup (Section~\ref{subsec:methods_prof_setup}) to provide a reference for the measurements using robot grippers.

\begin{figure}
    \centering
    \includegraphics[width=0.5\textwidth]{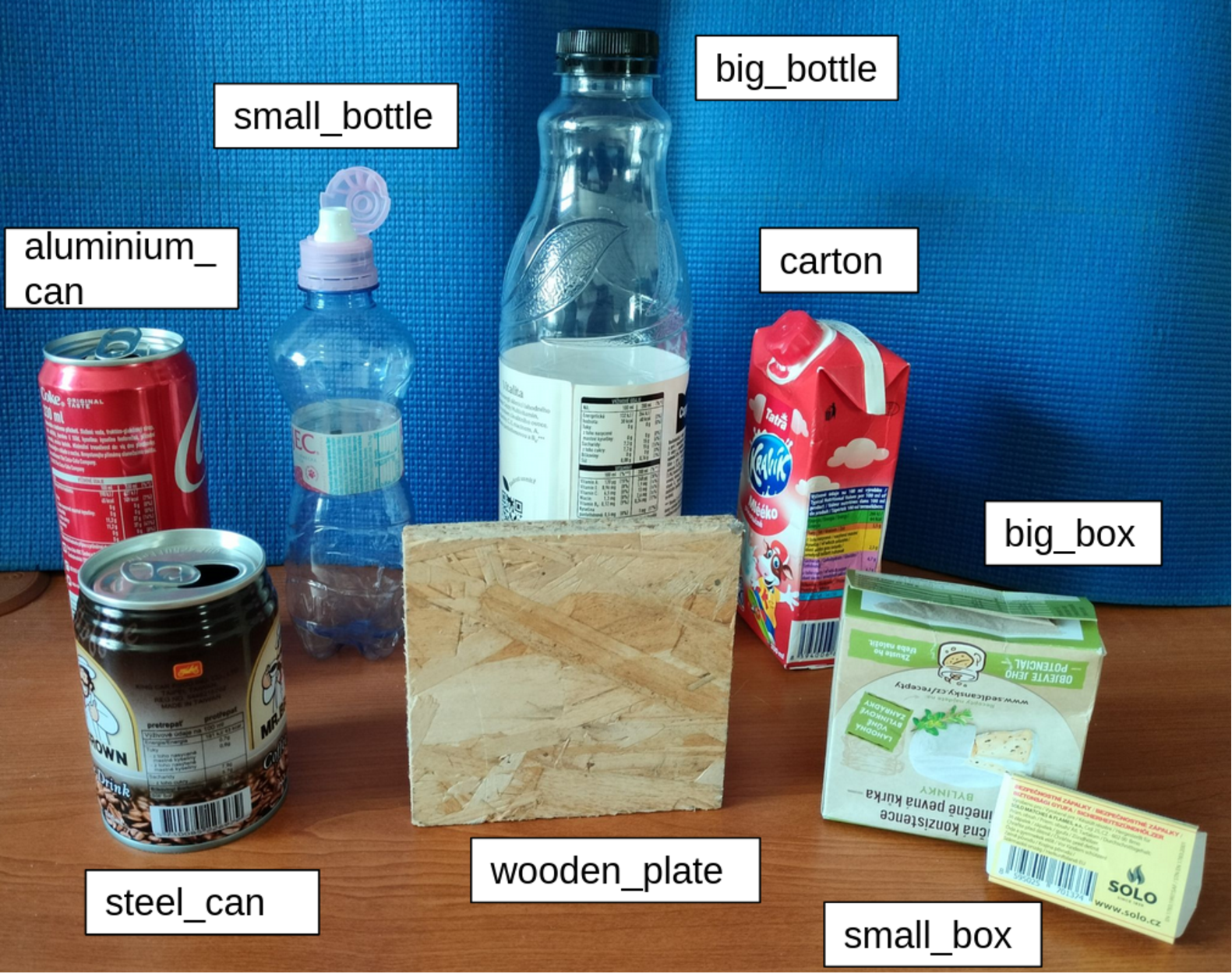}
    \caption{Waste sorting objects set.}
    \label{fig:recycling_objects_set}
\end{figure}

\paragraph{Waste sorting set}
The set shown in Figure~\ref{fig:recycling_objects_set} was acquired in the local grocery store to provide a simple test-bed resembling a single stream recycling plant where waste should be sorted according to the material composition into plastic, paper, metal, and other materials.

\subsection{Robot setups}
\label{subsec:methods_grippers}

For experimentation, the following robotic setups were used: the Universal Robot UR10 with the OnRobot RG6 gripper and the Kinova Gen3 robotic arm with the Robotiq 2F-85 gripper.
In addition, a force/torque sensor Robotiq FT300 was appended to the UR10 robot as a makeshift third setup.

\subsubsection{Robotiq 2F-85 gripper} 
\begin{figure}
    \includegraphics[width=0.48\textwidth]{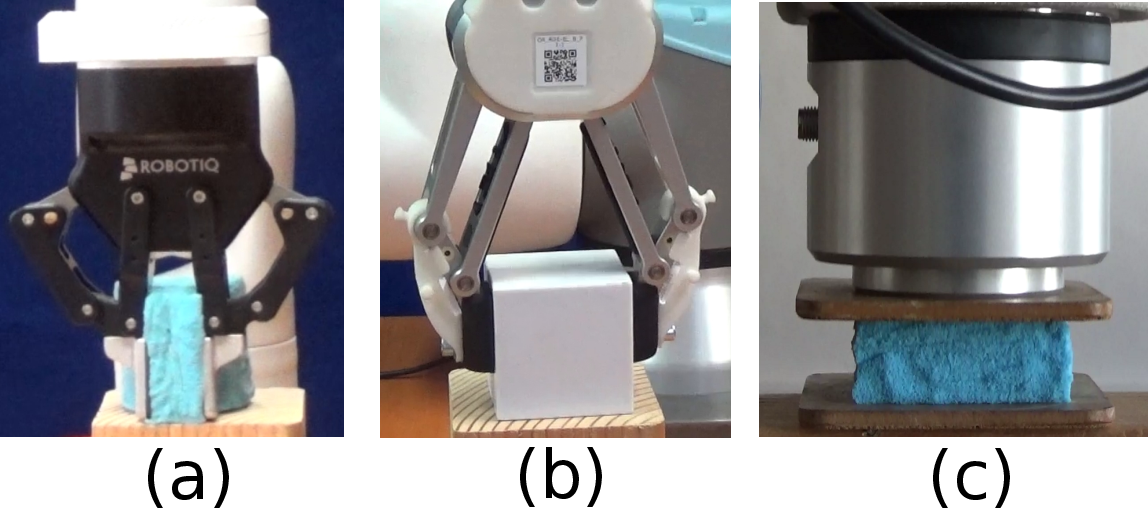}
    \caption{Robot devices. (a) Robotiq 2F-85 gripper; (b) OnRobot RG6 gripper; (c) Robotiq FT300 force/torque sensor.}
    \label{fig:grippers}
\end{figure}

This gripper (Fig.~\ref{fig:grippers}(a)) has 2 fingers with 85\,mm stroke and offers a grip force from 20\,N to 235\,N. The fingertips dimensions can be approximated to a rectangle of \begin{math}37.5\times22\,\mathrm{mm}\end{math}. Velocity control was used during object compression. This is quoted in \% of the maximum closing speed.

\textit{Action parameters}: Four different nominal squeezing speeds were executed: 0.68\% (approx. 1.6 mm/s), 14.45\% (30 mm/s), 50.85\% (80 mm/s), and 100\% (131.33 mm/s). These speeds gave rise to approximately 500, 300, 100, and 50 samples (points on the stress/strain curve) per compression cycle, respectively.

\textit{Sensory channels}: 2 channels: gripper position (gap between jaws 0--85 mm), motor current (A). 

\subsubsection{OnRobot RG6 gripper}
The gripper (Fig.~\ref{fig:grippers}(b)) has a 160\,mm stroke and adjustable force and gripper status feedback (digital and analog). Gripper fingertips' surface area is \begin{math} 866\,\mathrm{mm}^{2} \end{math}. Continuous closing while recording force feedback was not possible in our setup; it was approximated by incrementally increasing the force threshold.

\textit{Action parameters}: None. The gripper was commanded to close with 1 mm increments until a certain force threshold was reached. Then, the threshold was incrementally increased. The force range was 25-120 N, with 1 N increments. Approximately 100 data points were obtained during compression of a typical object.

\textit{Sensory channels}: gripper position (gap between jaws 0--160 mm) and flag for force threshold reached, plus the actual threshold (N). Data was collected only at the occasions when force threshold was reached.  

\subsubsection{Robotiq FT300 force/torque sensor}
The UR10e manipulator is equipped with a Robotiq FT300 force/torque sensor (Fig.~\ref{fig:grippers}(c)) that has a measuring range \(\pm 300\,\mathrm{N}\) and  \(\pm 30\,\mathrm{N}\cdot\mathrm{m}\) for every axis in Cartesian coordinate system. The sensor noise for the force measurements should be 0.033\,\%. 

\textit{Action parameters}: None. We pressed the objects with the robot flange against the table with a speed of 30 mm/s, which gave rise to approximately 30 samples on the stress/strain curve.  

\textit{Sensory channels}: position in z-axis (mm) and force in the downward direction (N).

\subsection{Gripper calibration}
\label{subsec:methods_gripper_calib}
We used a S9M/500 N force transducer (strain gauge) along with a ClipX amplifier system (both from HBM) to calibrate the force feedback from the robotic grippers.

The Robotiq 2F-85 gripper outputs motor current in the range 0-1 A. Measurements were collected by manually operating the gripper as well as using velocity control with the range of speeds used later in the experiments. The relationship of force and current was found to be nonlinear and the best fit obtained was: 
\begin{equation}
    F_{2F-85}=87.6i^3 - 216.0i^2 + 191.4i + 0.18
    \label{eqn:2F85_calibration}
\end{equation}
where \textit{i} was the current reading (in Amperes) obtained from the gripper. 
Additionally, the closing speeds in \% had to be calibrated and converted to mm/s (using the gripper position signal).

For the OnRobot RG6 gripper, feedback in Newtons is available. However, we have performed calibration as well and an approximately linear fit was obtained.
\begin{equation}
    F_{RG6-calib} = 0.8678 \times F_{RG6-uncalib} - 2.13
    \label{eqn:RG6_calibration}
\end{equation}

\subsection{Professional biaxial compression setup}
\label{subsec:methods_prof_setup}
\begin{figure}[htb]
    \centering
    \includegraphics[width=0.17\textwidth]{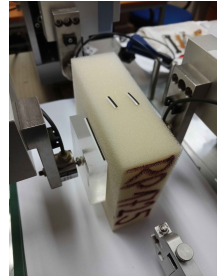}
    \caption{Professional setup for elasticity measurements.}
    \label{fig:prof_setup}
\end{figure}

This setup was deployed to provide reference or control measurements for the ``mixed set'' of objects (Section~\ref{subsec:methods_objects}).  
A biaxial experimental system Zwick/Roell was used (see Fig.~\ref{fig:prof_setup}). Samples were cyclically compressed between two antagonistic actuators, with a position resolution of $1\mu \cdot m$. The actuator axes were connected to force sensors U9B from HBM with a 0.5 kN range, set to $\pm 250$ N, suitable for static and dynamic force measurements (accuracy class 0.2; IP67). On every side, a transparent pad of plexiglass was attached, with dimensions $59\times59\times10$ mm and mass of 60 g. The sample was placed between the pads and cyclically compressed and decompressed from both sides. Six cycles of compression and decompression were performed for every sample. 

\textit{Action parameters}: Compression speeds: One sample (RP1725) was tested with 1, 1.6, 10 30, 50, 80 mm/s. Remaining samples from the mixed set were tested with 1.6, 30, 50, and 80 mm/s. Decompression speed was the same as compression speed with the exception of sample V4515 which took long to restore its volume---1.6 mm/s decompression was used. On average, we acquired 1200 points for the compression portion of each cycle. This is significantly higher than the data acquired from the robotic grippers.

\textit{Sensory channels}: Force (N) and displacement (mm) at every actuator.

\section{Experiments and results}
\label{sec:results}

The presentation of experiments starts with plots of mechanical response curves from the different devices in their standard units, followed by a transformation into stress/strain curves. The effect of compression speed and precycling is then analyzed. Several methods to estimate the Young's modulus of elasticity from the empirical data are compared in Section~\ref{subsec:results_young}, followed by estimations of viscoelasticity.
Finally, Section~\ref{subsec:single_stream_demo} presents a practical demonstration of the findings in a mock waste sorting scenario. Illustration of some of the experiments is available in the accompanying videos: Part 1, on the results of using different models to estimate elasticity and viscoelasticity \url{https://youtu.be/8DMMdEezC1M}; Part 2, showing the application of stiffness estimation to sort materials in a single-stream recycling scenario: \url{ https://youtu.be/XHiNKFZ158o}.

\subsection{Gripper effort and position}
\begin{figure*}
    \includegraphics[width=\textwidth]{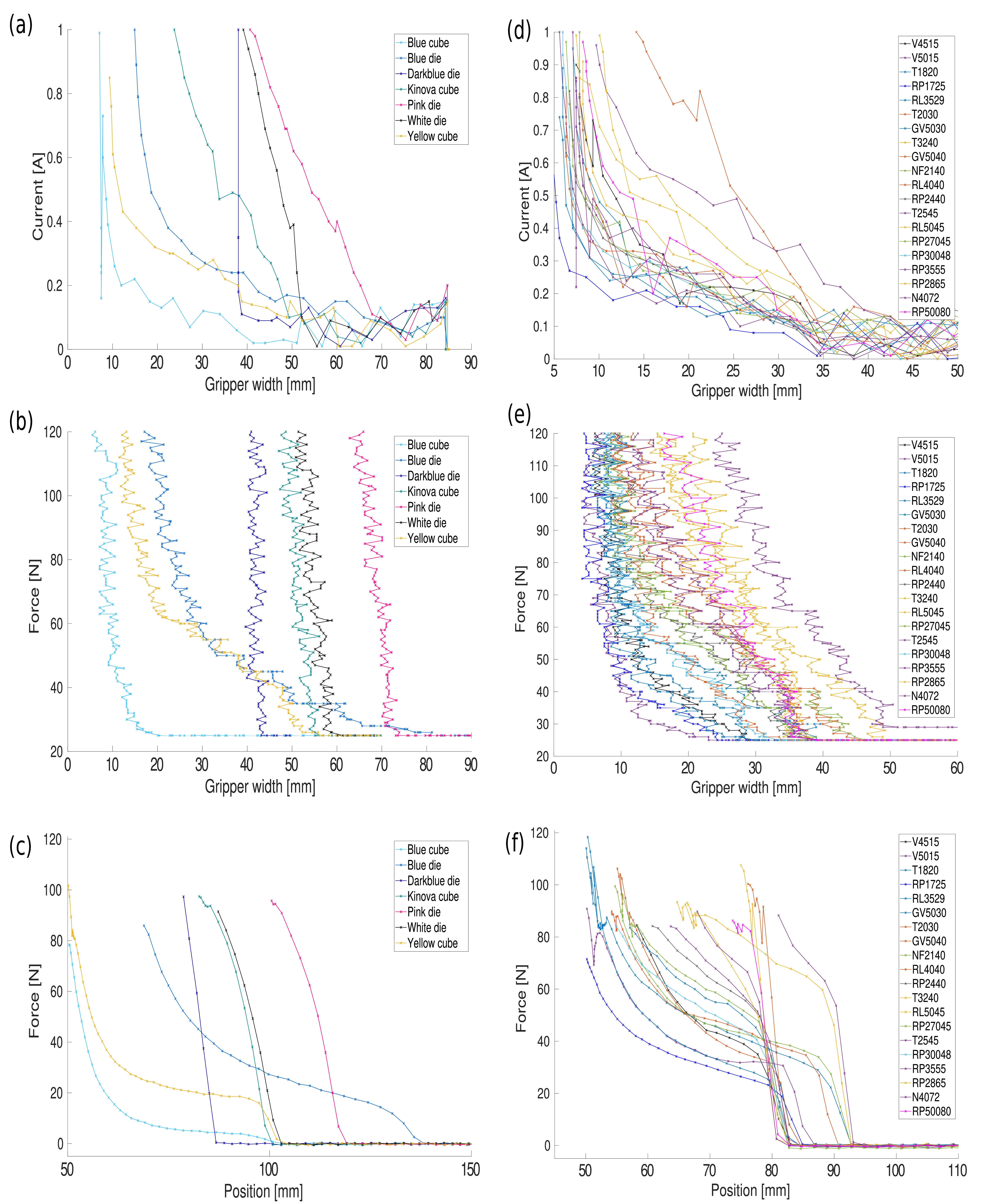}
    \caption{Raw gripper effort vs. position plots. \textbf{(a)-(c)} \textit{Cubes and dice set.} (a) Robotiq 2F-85 at $80 mm/s$; (b) OnRobot RG6; (c) Robotiq FT300 at $30 mm/s$. \textbf{(d)-(f)} \textit{Polyurethane foams set.} (d) Robotiq 2F-85 at $80 mm/s$; (e) OnRobot RG6; (f) Robotiq FT300 at $30 mm/s$.}
    \label{fig:raw_plots}
\end{figure*}

We compare the raw plots that can be obtained in the standard operation mode of every robotic device, as shown in Fig.~\ref{fig:raw_plots} for the cubes and dice.
No preprocessing or filtering of the data is applied. On the x-axes, object compression proceeds from right to left, as the gripper width decreases. Contact with the object is not detected and as the objects have different width, the curves corresponding to gripper effort start rising at different locations. 

For the Robotiq 2F-85 gripper, 50\% compression speed (approx. 79 mm/s) was used. Motor current is plotted against the gripper width. For the OnRobot RG6 gripper, force feedback was retrieved only during discontinuous object compression with incremental adjustment of the force threshold. During measurements with the FT300 sensor, the robot arm was commanded to press downward against the object with a speed of 30 mm/s.   

From Fig.~\ref{fig:raw_plots}, it is apparent that the mechanical response curves from all three devices have similar characteristics. The force/torque sensor (Fig.~\ref{fig:raw_plots}(c),(f)) delivers the most clean data. The lines on the OnRobot RG6 plot (Fig.~\ref{fig:raw_plots}(b),(e)) are noisy and not monotonic. This is because they were not collected during continuous compression of the objects, but through incremental increasing of the force threshold. The least accurate response curves come from the Robotiq 2F-85 (Fig.~\ref{fig:raw_plots}(a),(d)), mainly because there are too few data points (combination of limited sampling rate and high speed of compression).   

The 20 polyurethane foams constitute a more homogeneous object set, with similar dimensions and a narrower range of elasticity. Similar observations about the precision and continuity of data can be made for the deformable objects set and the polyurethane foams set.

\subsection{Stress/strain response curves}
To estimate material elasticity, the stress/strain relationship during compression of the material in question needs to be obtained (see Section~\ref{sec:physics_of_deformation} and Eqn.~(\ref{eq:Young})).  The key requirement is to detect the moment of contact of the gripper with the object, which marks the beginning of the stress/strain curve; the gripper width corresponds to $L_0$. To recognize the moment of contact, the force/effort needs to be measured, which is necessary at any rate to obtain stress. The accuracy of contact detection affects the results and is gripper dependent. Position can typically be measured accurately and thus obtaining strain as $\Delta L / L_0$ is straightforward. To obtain stress as Force/Area, the surface area will be the smaller from the gripper jaws and the side of the object. The crucial part, however, is the (compression) force. If force is not directly available from the gripper, its calibration is necessary (Section~\ref{subsec:methods_gripper_calib}).

\begin{figure*}
    \includegraphics[width=\textwidth]{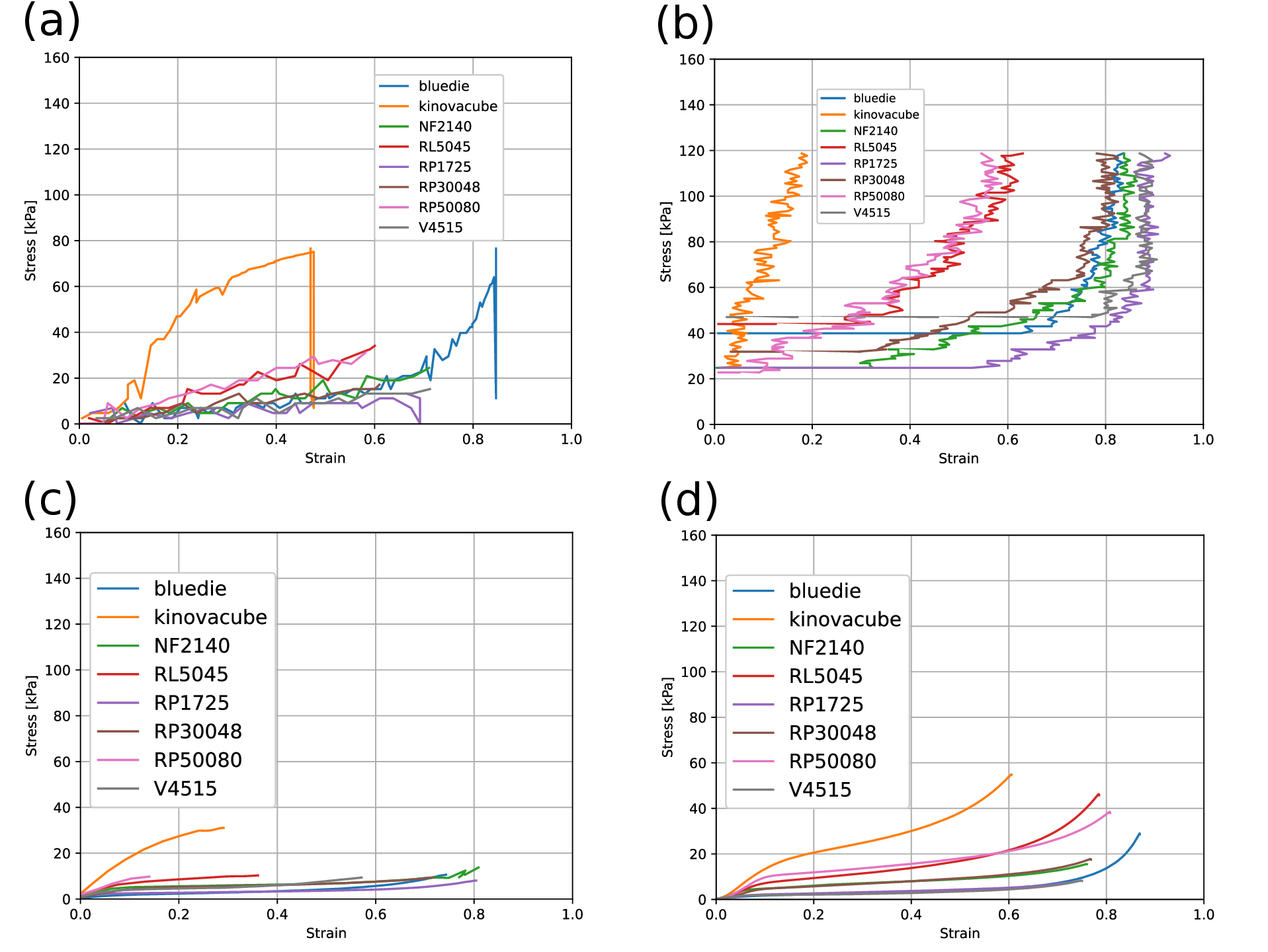}
    \caption{Mixed set -- stress/strain. (a)  Robotiq 2F-85 gripper. Compression speed 80 mm/s (50\%). (b) OnRobot RG6. Discontinuous compression. (c) Robotiq FT300 force/torque sensor. Compression speed 30 mm/s. (d) Professional setup. Compression speed 50 mm/s.}
    \label{fig:stress_strain_mixed}
\end{figure*}

After collecting raw compression data on the robotic setups from both the \textit{objects set} and the \textit{polyurethane foams set}, the \textit{mixed set} of foams and cubes was created (Section~\ref{subsec:methods_objects}) and tested on a professional biaxial measuring device (Section~\ref{subsec:methods_prof_setup}). The stress/strain curves are shown in Fig.~\ref{fig:stress_strain_mixed}. The curves from the professional setup, Fig.~\ref{fig:stress_strain_mixed}(d), will serve as reference. We observe that the samples are clustered with the ``Kinova cube'' being the least elastic material, followed by RL5045 and RP50080 foams. The rest of the samples occupy the bottom of the plot. This clustering is apparent also on the data from the RG6 gripper, Fig.~\ref{fig:stress_strain_mixed}(b), and the FT300, Fig.~\ref{fig:stress_strain_mixed}(c); it is less pronounced for the 2F-85 in Fig.~\ref{fig:stress_strain_mixed}(a). The nonlinear response visible in the feedback recorded by the professional setup is partially preserved in the plots from the other devices. An important observation is the range on the y-axis. The absolute values of stress differ significantly, which limits the possibilities of estimation of the real value of the modulus of elasticity. The 2F-85 values seem most in line with the professional setup; the FT300 is underestimating the stress by approximately a factor of 2; the RG6, on the other hand, is overestimating the stress by the same factor. For estimation of real values of elastic moduli, such discrepancy is problematic. 

\subsection{Effect of compression speed and precycling}

\begin{figure*}[htb!]
    \includegraphics[width=\textwidth]{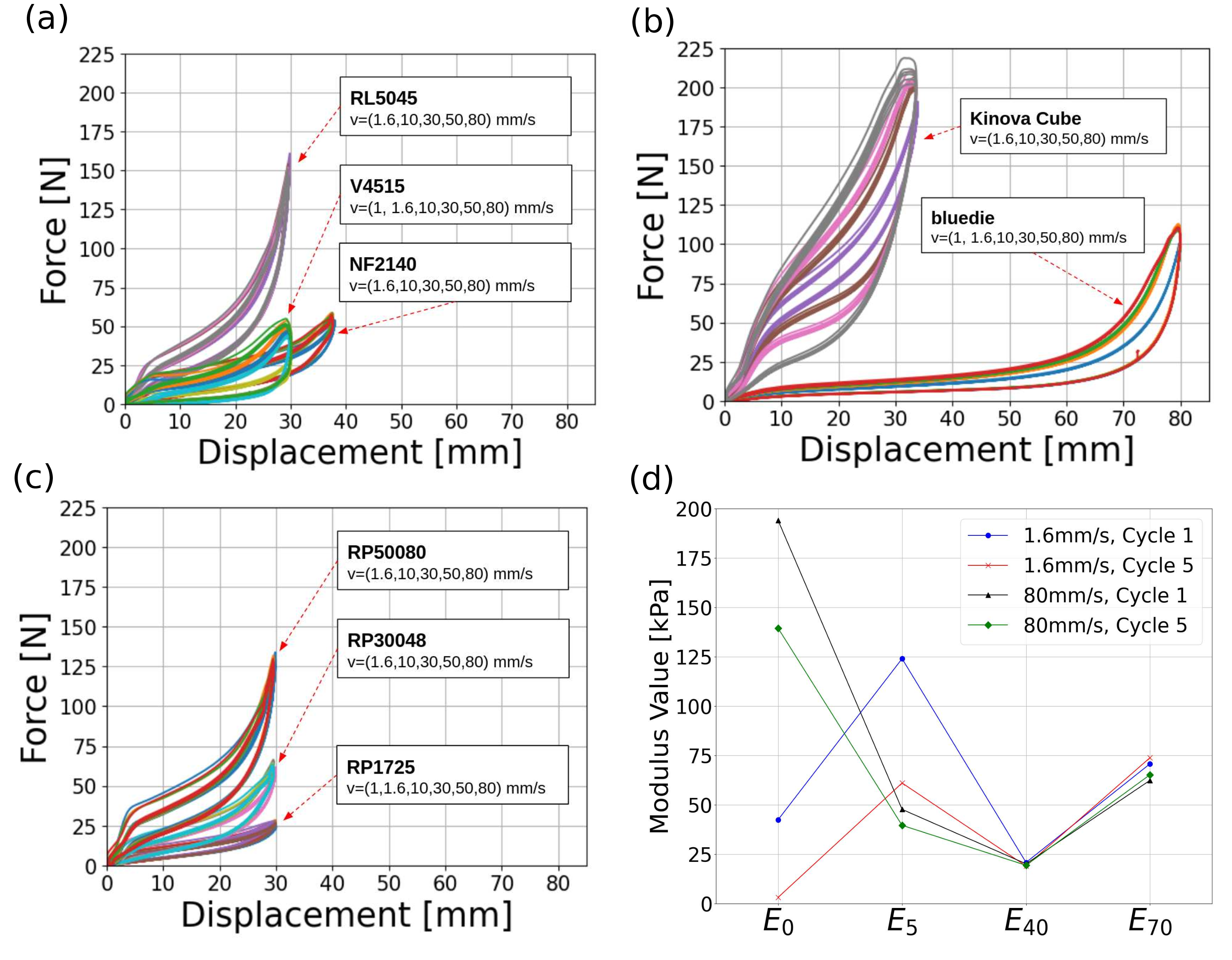}
    \caption{Compression and decompression cycles with different speeds on professional device -- Mixed set.  (a)-(c) Visualizations from biaxial compression device -- Force at one sensor over displacement of one jaw. Six compression and decompression cycles for every speed and sample. (d) Variation of estimated moduli with cycle number and speed, for the foam RP50080. 
    }
    \label{fig:profsetup_speeds_cycles}
\end{figure*}

The ISO standard for low-density polymeric materials \cite{iso3386polymeric} prescribes that samples be compressed at a speed of 1.67 mm/s and that the stress/strain characteristics should be evaluated during the 4th compression cycle. To assess the effect of these parameters, we first used the biaxial professional setup. The compression and, in this case, also decompression  curves for the objects of the mixed set are shown in Fig.~\ref{fig:profsetup_speeds_cycles}, (a)-(c). Note that the upper part of every curve corresponds to compression. With higher speeds, the difference between the compression and decompression curve (hysteresis) is bigger. Individual compression cycles (1-6) did not have a significant effect on the overall shape. Fig.~\ref{fig:profsetup_speeds_cycles}(d) shows how the estimated modulus varies at different values of strain and with compression speed and cycle.

We have further studied the effect of speed and precycling on estimating Young's modulus (Section~\ref{subsubsec:young_all_devices} and Fig.~\ref{fig:young40_all_devices}) across the different devices. Precycling indeed did not significantly affect the results. Slower compression speeds tend to have more samples per compression cycle, which is advantageous when using regression models to estimate material properties from feedback.

\subsection{Young's modulus of elasticity}
\label{subsec:results_young}
For the materials we tested, the stress/strain curves are nonlinear. Therefore, the modulus of elasticity will be different at different points on the curve. Thus, we estimate at different values of strain (0\%, 5\%, 40\%, and 70\%) by locally estimating the gradient, and a ``linear'' modulus that assumes the whole curve is linear. 

\subsubsection{Positions on stress/strain curve and window size}
The response curves obtained from grippers are not densely sampled and are noisy.
Hence, prior to modulus estimation, the feedback is filtered using the Savitzky-Golay filter which smooths data using least-squares polynomial fitting on smaller window sizes. 
For the analysis of the data from the OnRobot RG6 and the Robotiq 2F-85 in this section, calibrated force values (Section~\ref{subsec:methods_gripper_calib}) are used. 

\begin{figure*}[!htb]
    \includegraphics[width=\textwidth]{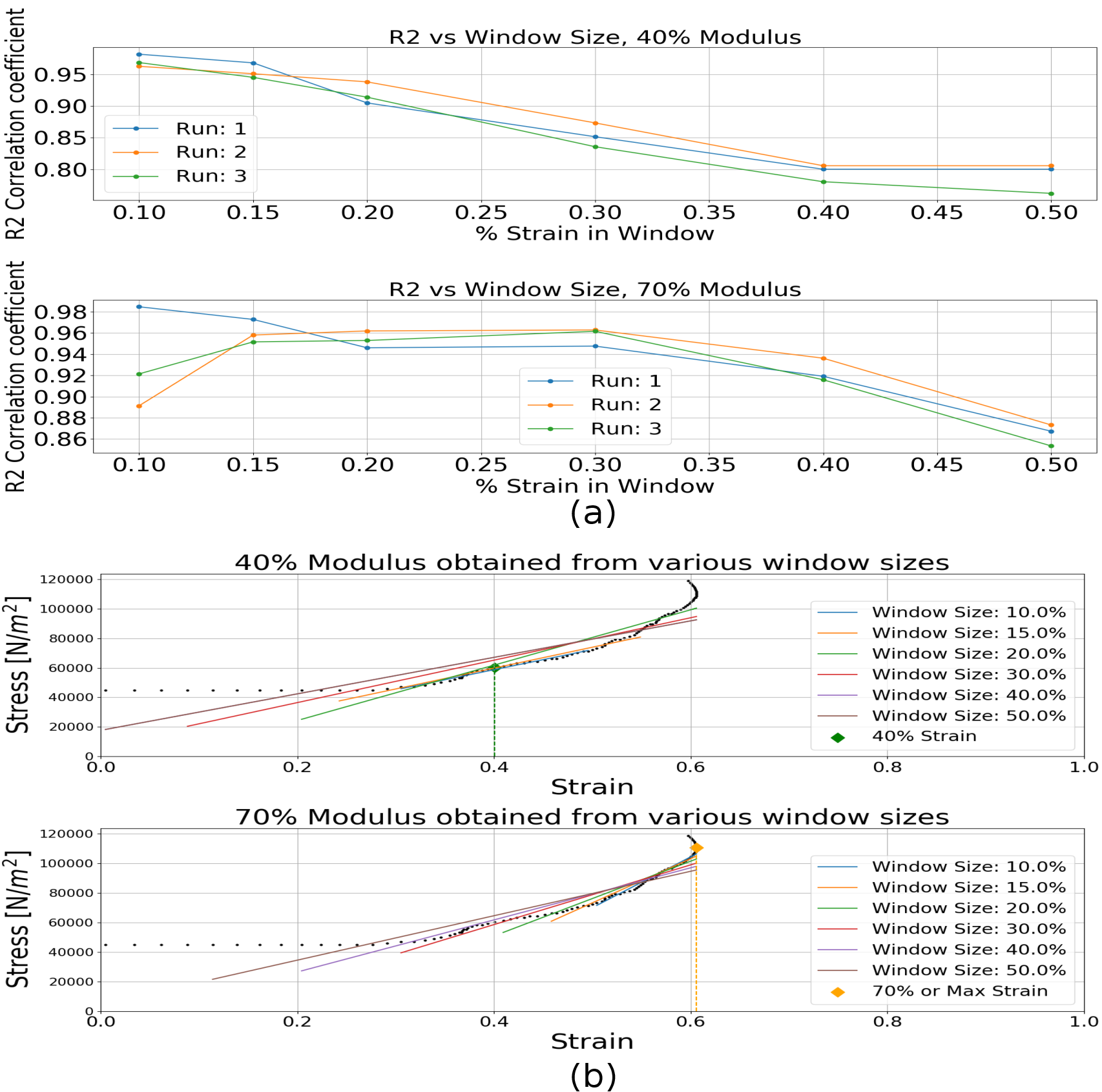}
    \caption{Young's modulus estimation -- OnRobot RG6 gripper. (a) Variation of $R^{2}$ scores with window size. (b) The best fit line plotted for different window sizes. Four cycle 5 of compression of the foam RL5045. Similar curves were obtained for all foams.}
    \label{fig:window_size_experiments}
\end{figure*}

To find the modulus at any chosen strain percentage, we create a neighborhood around the strain value and find the slope of the best fit line in that neighborhood using \textit{scikit-learn}'s \textit{LinearRegression} module. To ensure that the best fit line obtained is actually representative of the slope at the chosen point, we vary window sizes to obtain the highest $R^{2}$ correlation as a measure. It is observed that this $R^{2}$ vs. window size variation is consistent in shape for the modulus values obtained at 40\% and 70\% strains. As shown in Fig.~\ref{fig:window_size_experiments}(a), it can be seen that smaller window sizes have the highest $R^{2}$ scores at 40\% and 70\% strain. Alternatively, the $R^{2}$ scores for the best fit lines near the 0\% and 5\% strain values are very poor. 

The window size for our best fit lines was fixed at $\pm$ 10\% strain on either side. All results obtained from the robotic grippers are based on this window size.

\subsubsection{Estimating Young's modulus from professional setup}
We calculate the modulus values from the data from the professional setup. Modulus at 0\%, 5\%, 40\%, and 70\% was estimated using linear regression. Also, a ``linear'' modulus was calculated for the entire curve. For one sample, RP50080, statistical tests (t-test) were conducted to assess whether the compression cycle or the speed affect the modulus estimation. For the moduli at 0\% ($E_0$) and 5\% ($E_{05}$), significant differences between cycle 1 and cycle 5 (p=0.01) were found. For $E_{40}$ and $E_{70}$, no significant differences were found. Statistical tests comparing moduli estimation between slow speeds (1-10 mm/s) and fast speeds (30-50 mm/s) were also conducted and no significant differences were found. See also Fig.~\ref{fig:profsetup_speeds_cycles}. 

The moduli $E_{40}$ and $E_{70}$ moduli for the mixed set of samples are shown in Table~\ref{tab:prof_setup_moduli} (mean over compression speeds, first compression cycle only). The error is presented as a ratio between the standard deviation and the mean. Since the sampling was dense, approximately 2\% window on both sides of the desired strain value was sufficient. 

\begin{table}[!htb]
    \caption{Mean values of 40 \% modulus ($E_{40}$) and 70 \% Modulus ($E_{70}$) for mixed object set on the professional setup; first compression cycle, all speeds.}
    \label{tab:prof_setup_moduli}
    \begin{tabular}{@{}lll@{}}
    \toprule
        Foam & $E_{40}$ [kPa] (Err) & $E_{70}$ [kPa] (Err)\\
    \midrule
        Blue die & 6.783 (10.0\%) & 45.609 (15.2\%)\\
        Kinova cube & 82.638 (17.9\%) & 82.638 (17.9\%)\\
        NF2140 & 9.756 (10.2\%) & 46.047 (18.6\%)\\
        RL5045 & 26.323 (4.5\%) & 116.240 (14.5\%)\\
        RP1725 & 6.391 (8.6\%) & 20.557 (29.0\%)\\
        RP30048 & 12.439 (5.7\%) & 50.479 (20.6\%)\\
        RP50080 & 20.044 (3.53\%) & 75.912 (9.9\%)\\
        V4515 & 9.952 (27.32\%) & 42.400 (30.68\%)\\
    \bottomrule
    \end{tabular}
\end{table}

\subsubsection{Young's Modulus -- all devices}
\label{subsubsec:young_all_devices}

Data from all the different devices and different settings were used for Young modulus estimation on the mixed set sample. Estimations at 0\% and 5\% strain are not reported here as these were hard to obtain in all but the professional setup due to sparse sampling and presence of noise. At least five compression and decompression cycles were performed with every setting and the moduli were estimated at cycle 1 and cycle 5. Speed was varied for the professional setup and for the Robotiq 2F-85 gripper. In addition, to account for the effect of the surface area of the gripper jaws, the same pads (59 x 59 mm) used in the professional setup were mounted onto the OnRobot RG6 and Robotiq 2F-85 jaws as an additional setting. 

\begin{figure}[!htb]
    \includegraphics[width=0.49\textwidth]{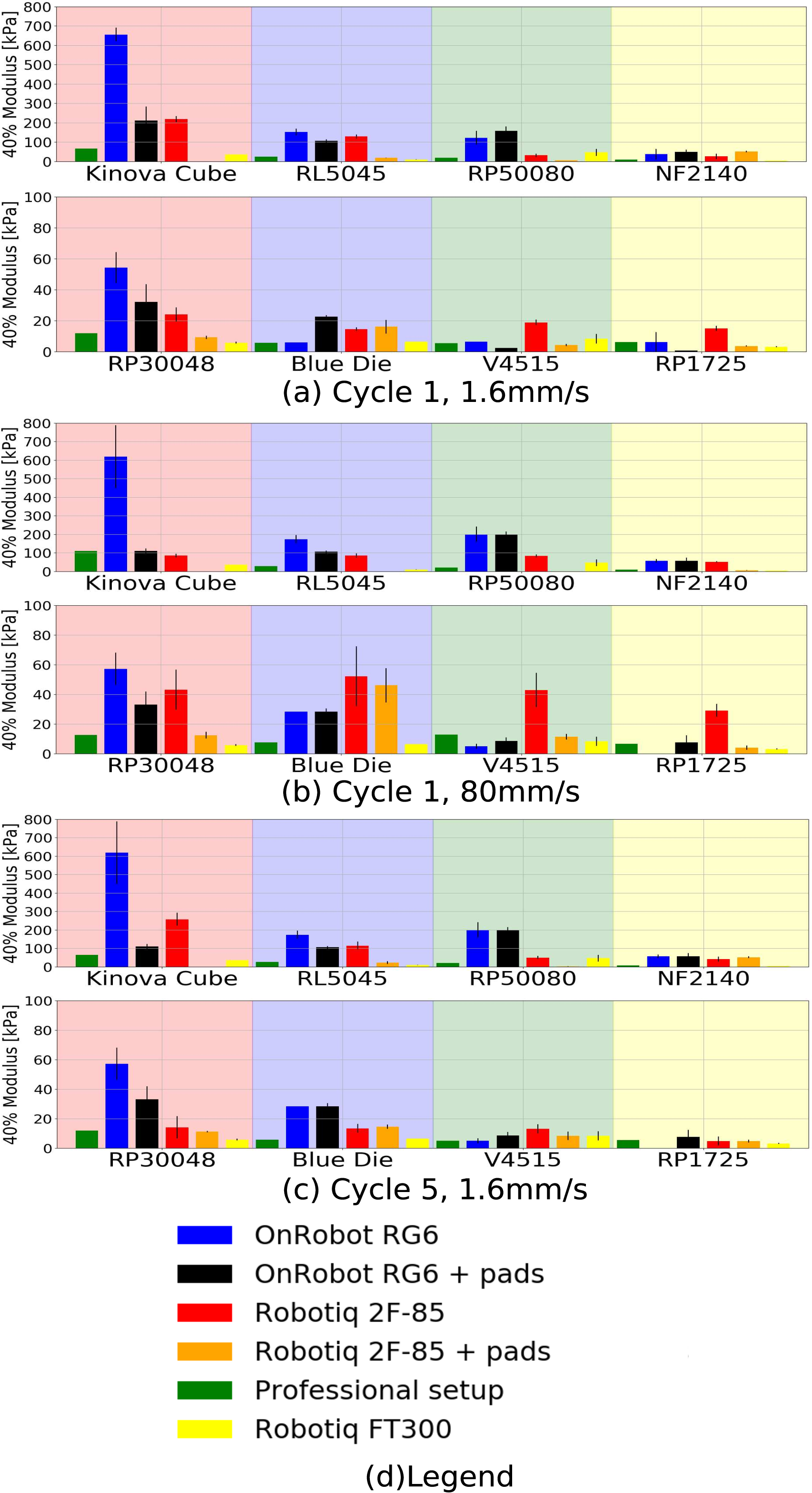}
    \caption{Mixed set -- Young modulus at 40\% strain. Estimation using different devices and settings.}
    \label{fig:young40_all_devices}
\end{figure}

The results of Young modulus estimation at 40\% strain are shown in Fig.~\ref{fig:young40_all_devices}. The green bars in every subplot are from the professional setup and serve as reference. The following observations can be made: (1) no robotic device provides an accurate and reliable estimate of Young's moduli corresponding to the values obtained from the professional setup;
(2) adding the pads and thus increasing the surface area for the 2-finger grippers brings the absolute values closer to the reference, but adds noise to the estimation (disrupting the relative ordering of samples by elasticity);
(3) the FT300 force sensor provides the most stable estimation, albeit the absolute values of Young's modulus are significantly lower compared to the reference. This is consistent with the raw strain/stress curves in Fig.~\ref{fig:stress_strain_mixed}; 
(4) there is no clear effect of precycling (5th versus 1st compression cycle); 
(5) bigger compression speed degrades the estimation for the Robotiq 2F-85 gripper.

\subsubsection{Linear modulus and relative elasticity estimation}
\label{subsubsec:results_linear_modulus}
\begin{figure*}[!htb]
    \includegraphics[width=0.95\textwidth]{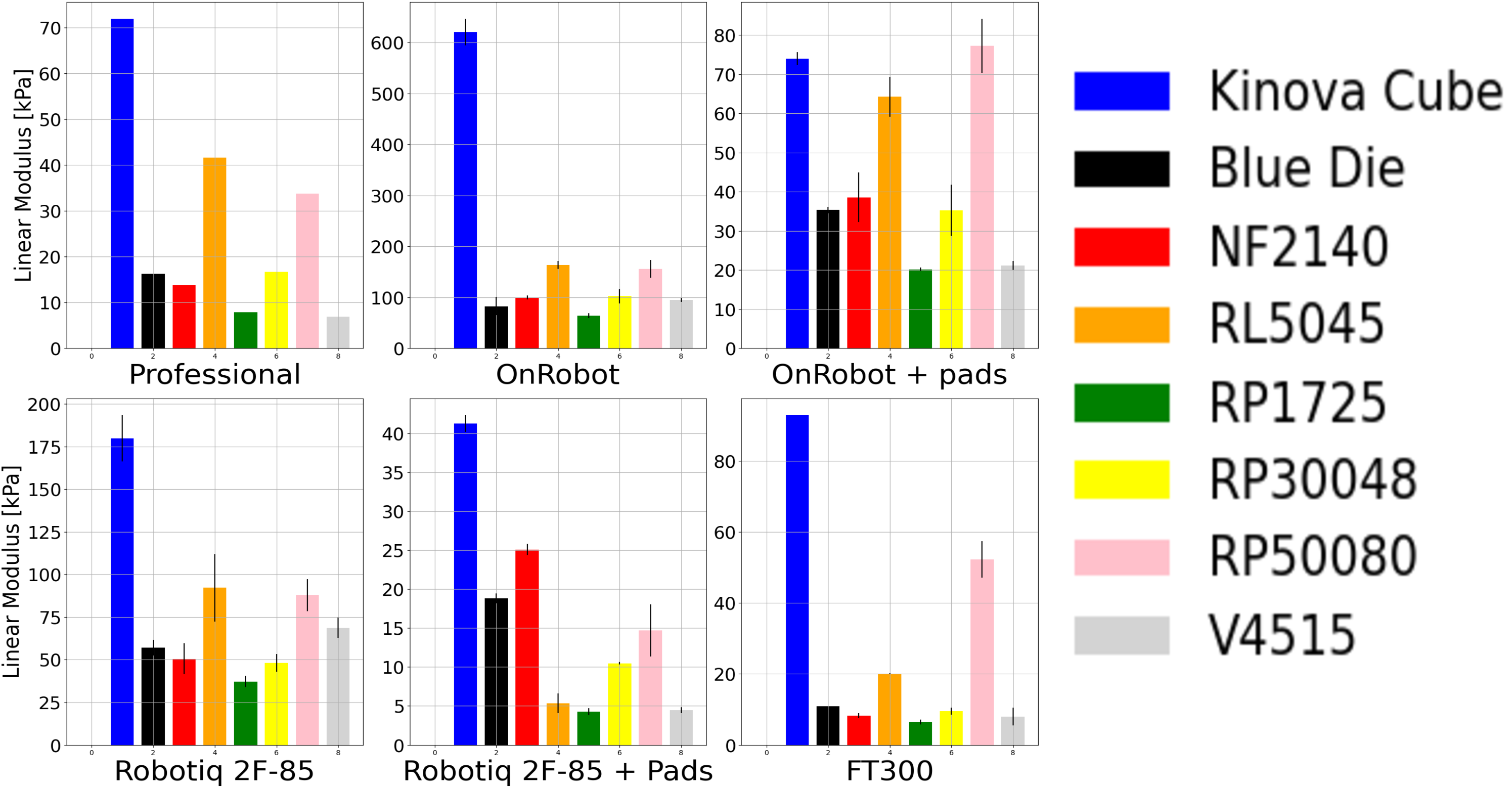}
    \caption{Linear modulus obtained by different grippers (compression speed = 1.6mm/s); Mixed set. Please note the different scale on the y-axes.}
    \label{fig:tot_mod}
\end{figure*}

The previous sections have shown that the capability of robotic grippers to deliver accurate estimates of Young modulus at different strain values are limited. Due to noise and limited resolution of the devices, local windows for estimation at e.g. 40\% and 70\% are unreliable. From a practical point of view, relative discrimination of objects based on their elasticity will often suffice. After experimentation, we propose to measure the stress/strain ratio over the whole compression area. This assumption is not correct since the mechanical response curves are clearly nonlinear, but it provides stable results in terms of ordering the samples by their elasticity, largely consistent across the different devices -- see Fig.~\ref{fig:tot_mod} and below:  
\begin{multline}
Kinova Cube > RP50080, RL5045 > V4515,\\ RP30048, NF2140 > RP1725, Blue Die    
\end{multline}

\subsection{Viscoelasticity}
\label{subsec:results_viscoel}

\subsubsection{Viscoelasticity and the hysteresis loop}
\begin{figure}[!htb]
    \centering
    \includegraphics[width=0.48\textwidth]{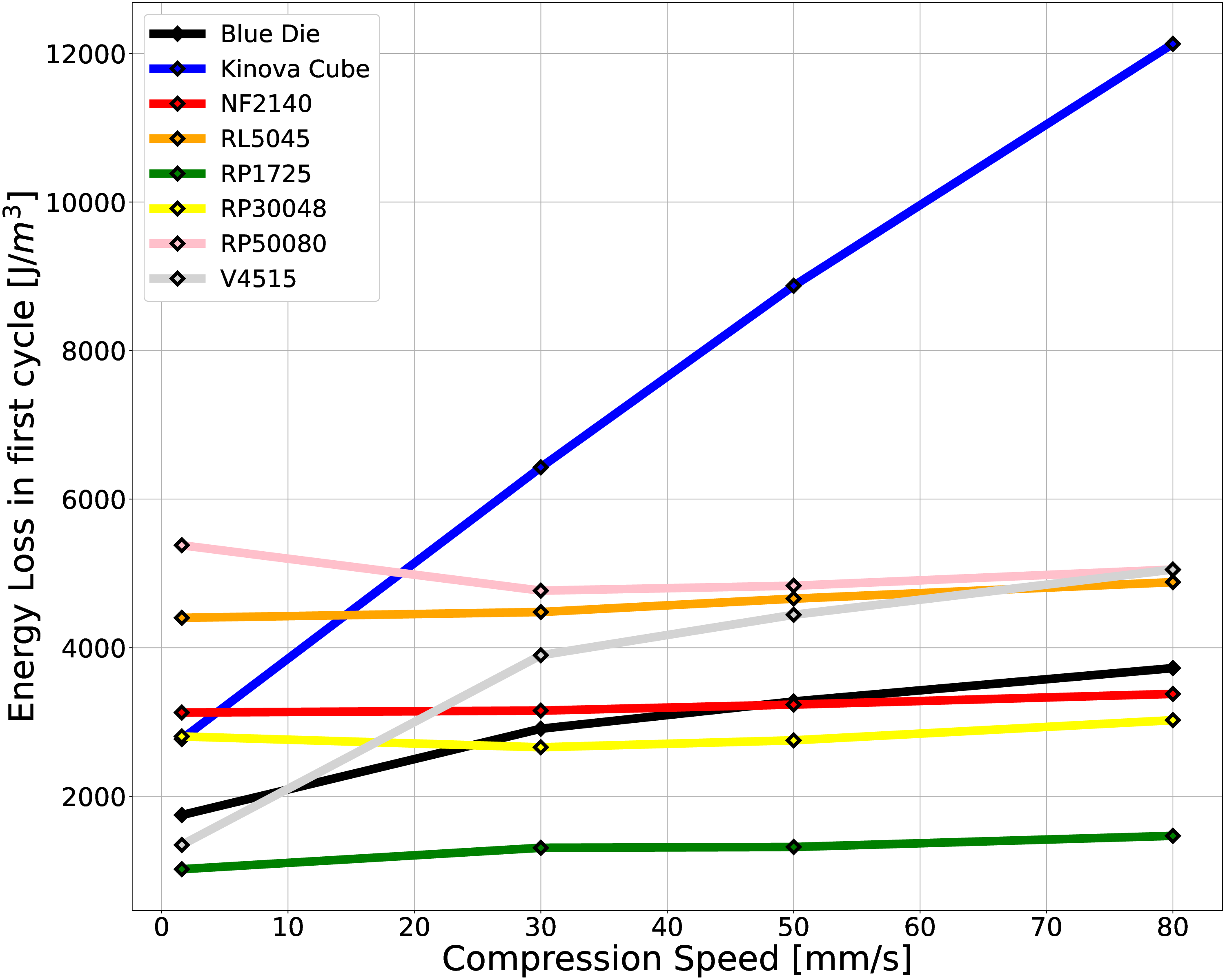}
    \caption{Viscoelasticity -- Mixed set, on the professional setup. Variation of the amount of energy lost per compression and release cycle with different speeds.}
    \label{fig:comp_coeff}
\end{figure}

All the materials composing the mixed set have hysteresis loops (explained in Section~\ref{subsubsec:Hysteresis_Area}) as can be seen in Fig.~\ref{fig:profsetup_speeds_cycles}.   
The amount of energy lost is shown to be linearly proportional the compression speed in Fig.~\ref{fig:comp_coeff}. The linear coefficient of the energy lost vs. speed graph is a characteristic of the material, and we label it $\eta$. The values of this energy loss coefficient obtained from the professional setup are listed for all materials in Table~\ref{tab:ISO_hys_vals}. 
The materials fall approximately into two groups based on the values of $\eta$. The viscous component is pronounced for the Blue die, Kinova cube, and the V4515 polyurethane foam. The Blue die and Kinova cube differ from the foams in that they are not completely homogeneous---they have a smooth outer shell that encloses the inside of the material and changes the properties. The difference in viscosity of the polyurethane foams may have to do with their porosity.  
Viscosity can provide and additional dimension next to elasticity to improve discrimination of soft materials.

\begin{table}[!htb]
    \caption{Viscoelastic coefficient obtained by calculating the hysteresis energy loss for multiple speeds on the professional setup.}
    \label{tab:ISO_hys_vals}
    \begin{tabular}{@{}ll@{}}
    \toprule
        Foam & $\eta$ (x $10^{3}$)[N.s.$m^{-2}$] \\
    \midrule
        Blue die & 24.717 \\
        Kinova cube & 119.487 \\
        NF2140 & 3.267 \\
        RL5045 & 6.314 \\
        RP1725 & 5.392 \\
        RP30048 & 2.960 \\
        V4515 & 45.812 \\
    \bottomrule
    \end{tabular}
\end{table}

\subsubsection{Elasticity and viscoelasticity -- Kelvin-Voigt and Hunt-Crossley models}
\label{subsec:results_viscosity_2f85}
All the models to characterize the viscous characteristics of the samples introduced in Section~\ref{subsec:DynMod_Hyst}---area inside the hysteresis loop, Kelvin-Voigt model, and Hunt-Crossley model---were applied to the mixed set samples as measured by the Professional setup and the Robotiq 2F-85 gripper.\footnote{Note that the OnRobot RG6 gripper cannot be used since continuous object compression with different speeds while recording force feedback is not possible.} Both devices were used to compress and release the samples with the following speeds: 1.6, 30, 50, 80 mm/s.  
To find parameters of the Kelvin-Voigt model and the Hunt-Crossley model, data from the first and fifth cycles for different speeds was fit into Eqn.~(\ref{eqn:KVModel}) and Eqn.~(\ref{eqn:HCModel}) individually and then averaged.

\begin{figure*}
    \includegraphics[width=\textwidth]{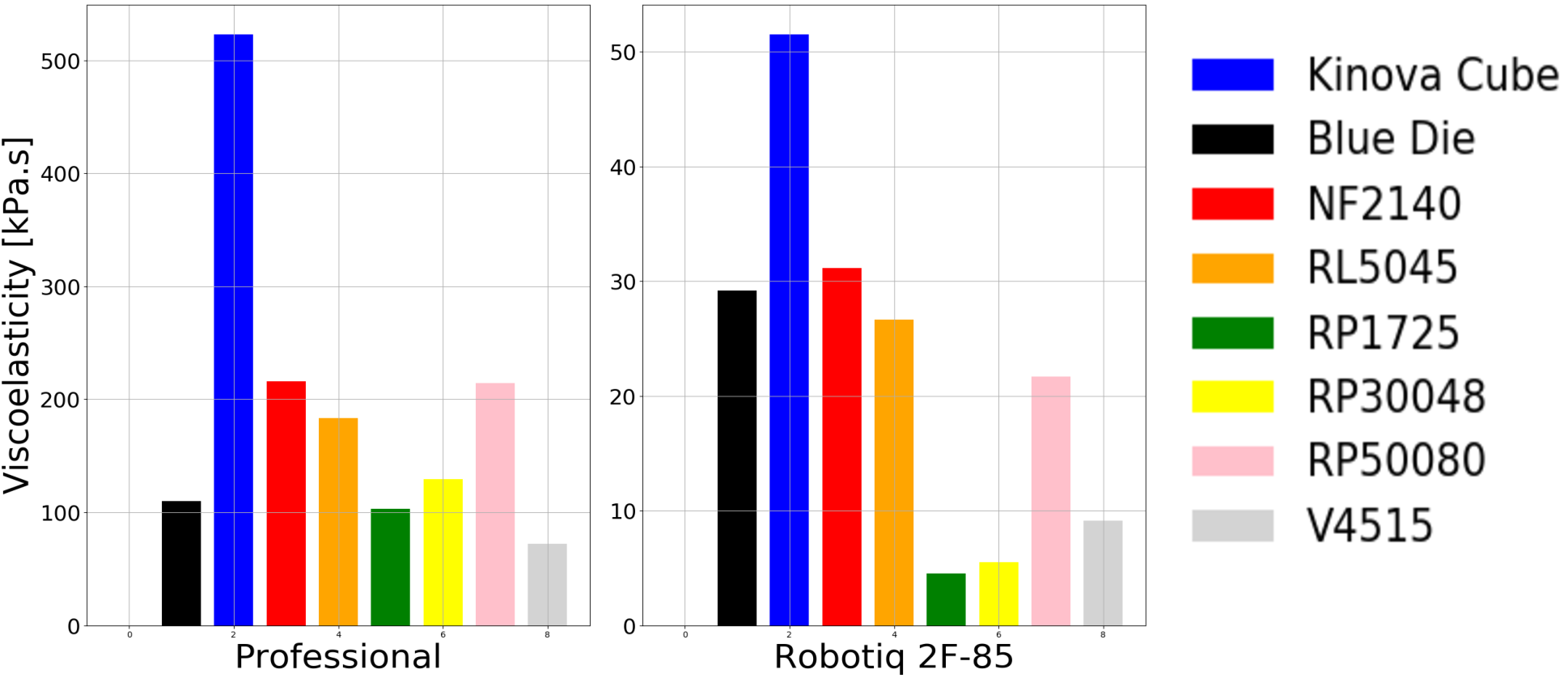}
    \caption{Comparing viscoelasticity ($\eta$) values obtained from the Professional Setup (left) and the Robotiq 2F-85 gripper (right), averaged over all testing speeds: 1.6, 30, 50 and $80 mm/s$.}
    \label{fig:VE_results}
\end{figure*}

An overview of the results is in Fig.~\ref{fig:VE_results}. The correspondence between the viscosity/damping parameter obtained from the professional setup and from the Robotiq 2F-85 gripper is not satisfactory for the area inside the hysteresis loop ($R^{2}=0.46$) and the Kelvin-Voigt model ($R^{2}=0.35$) is not satisfactory. Only the Hunt-Crossley model ($R^{2}=0.81$) provides a reasonable correlation---we will thus use this model in what follows.

\subsection{Material discrimination from elasticity and viscoelasticity}
\begin{figure}[!htb]
    \centering
    \includegraphics[width=0.5\textwidth]{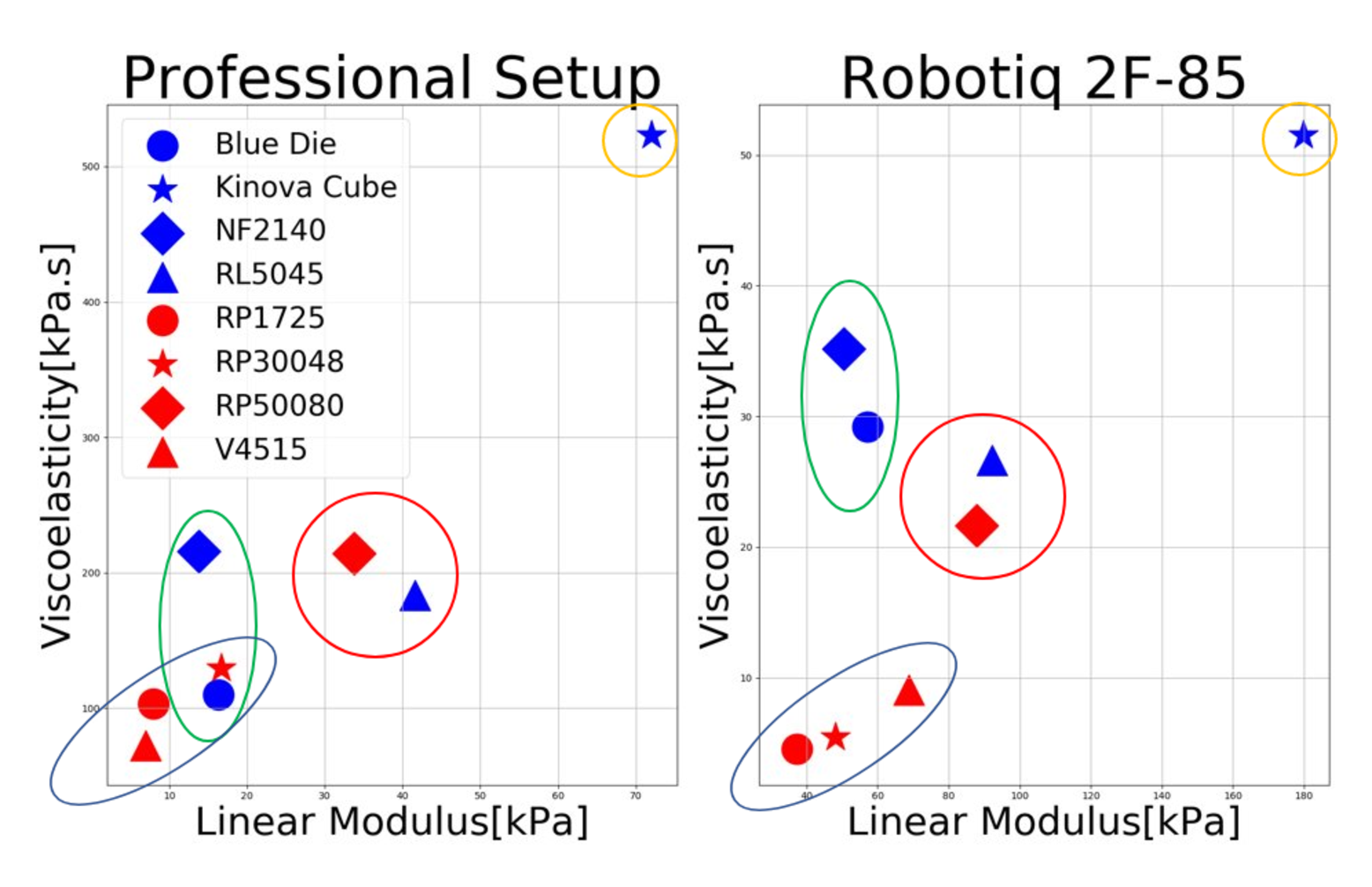}
    \caption{2D object discrimination -- elasticity using the linear assumption (Sec.~\ref{subsubsec:results_linear_modulus}) on x-axis and viscosity as obtained from the Hunt-Crossley model ($\eta$ Eqn.~\ref{eqn:HCModel}) on y-axis. (Left) Professional setup. (Right) Robotiq 2F-85. Clusters formed are marked by hand.}
    \label{fig:2D_grid}
\end{figure}

Given the limited capabilities of robotic grippers to gauge material elasticity (Sec.~\ref{subsec:results_young}) and viscosity (Sec.~\ref{subsec:results_viscosity_2f85}) alone, if the goal is online discrimination of deformable objects, it is advantageous to use both these quantities simultaneously. This is visualized in Fig.~\ref{fig:2D_grid} with the values obtained from the professional setup and the Robotiq 2F-85 gripper side by side. One can observe that there is a good qualitative match between the two measuring devices and that the second dimension importantly aids the possibility of discriminating the objects.

\subsection{Online waste sorting demonstrator}
\label{subsec:single_stream_demo}
Finally, we demonstrate the feasibility of object classification via physical property estimation in a mock waster sorting scenario. A robotic arm was presented with a sequence of objects that it had to grasp and lift, classify into one of four classes of recyclable materials, and deposit them in their respective bins. We demonstrated this on the Kinova Gen3 robot with the Robotiq 2F-85 gripper. A video is available at \url{https://youtu.be/XHiNKFZ158o}.

A new dataset with the objects shown in (Fig.~\ref{fig:recycling_objects_set}) was collected. These objects were typical and potentially recyclable grocery items. The aim was to sort these objects into different classes of recyclable materials, utilizing the Hunt-Crossley model whose performance we found to be the best. The gripper was provided with the grasping position and initial thickness of the object as input, and estimated the stiffness and damping values of the object from the feedback from single squeezing action. Stiffness and damping are properties that represent the elasticity and viscoelasticity of objects, while taking into account not only the material of the object, but also the shape and geometry. Our four material categories were \textit{Paper and Cardboard}, \textit{PET and Plastic}, \textit{Sheet Metal Containers} and \textit{Too Stiff}. First, sample data was collected from the set of recyclable objects on the professional setup and on the Kinova Gen3 robot, and stiffness ranges were derived for each of these categories. Then we set up a routine that repetitively explored objects and sorted them based on estimated stiffness.

\begin{figure*}[!htb]
    \centering
    \includegraphics[width=\textwidth]{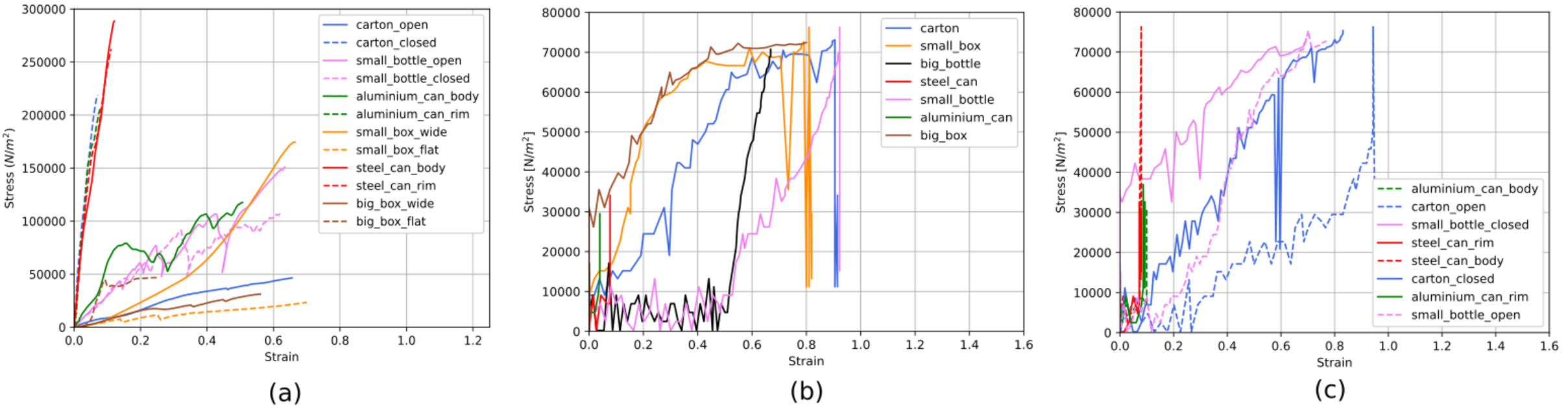}
    \caption{Raw stress-strain plots for the recycling objects set. (a) Using the professional setup, compression at $1.6 mm/s$. (b) Using the Kinova Gen3 arm with the Robotiq 2F-85 gripper, with different objects, compression at $1.6 mm/s$. (c) Using the robot with different grasps on the same objects, compression at $1.6 mm/s$.}
    \label{fig:recycling_set_raw}
\end{figure*}

\begin{table}[!htb]
    \caption{Hunt-Crossley parameter estimates -- Robotiq 2F-85 gripper (at 1.6 mm/s), Waste sorting set.}
    \label{tab:recycling_set_values}
    \begin{tabular}{@{}llll@{}}
    \toprule
        Object & Material & K[$N/m^2$] & $\eta$[$Pa.s$] \\
    \midrule
        small\_box & Cardboard & 15569 & 26466 \\
        big\_box & Cardboard & 13106 & 29013 \\
        carton & PET & 16893 & 28056 \\
        small\_bottle & Plastic & 17711 & 1278 \\
        big\_bottle & Plastic & 19898 & 1979 \\
        aluminium\_can & Sheet Metal & 35705 & 88685 \\
        steel\_can & Sheet Metal & 44303 & 35462 \\
    \midrule
        Object & n & $R^2$ & \\
    \midrule
        small\_box & 0.46 & 0.90 & \\
        big\_box & 0.30 & 0.85 & \\
        carton & 0.77 & 0.86 & \\
        small\_bottle & 2.55 & 0.98 & \\
        big\_bottle & 5.40 & 0.86 & \\
        aluminium\_can & 0.48 & 0.56 & \\
        steel\_can & 0.60 & 0.70 & \\
    \bottomrule
    \end{tabular}
\end{table}

\begin{figure}[!htb]
    \centering
    \includegraphics[width=0.5\textwidth]{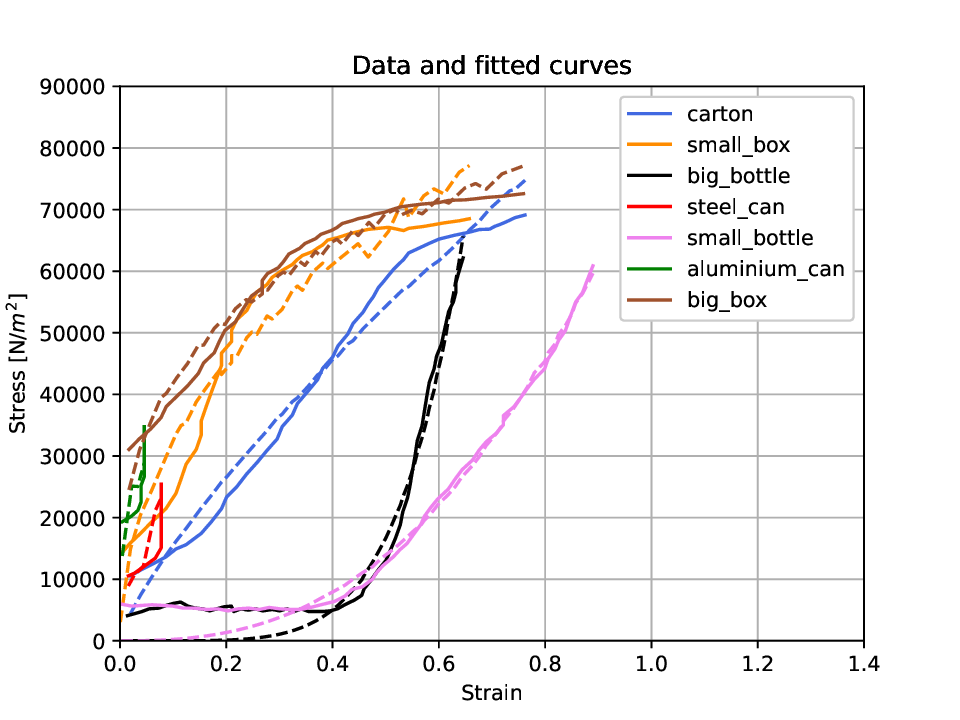}
    \caption{Model estimation performance. The solid lines represent actual data from the objects on the Robotiq gripper compressing the objects at $1.6 mm/s$. The dashed lines are recreated by the Hunt-Crossley model.}
    \label{fig:estimation_recreation}
\end{figure}

\begin{figure*}[!htb]
    \centering
    \includegraphics[width=\textwidth]{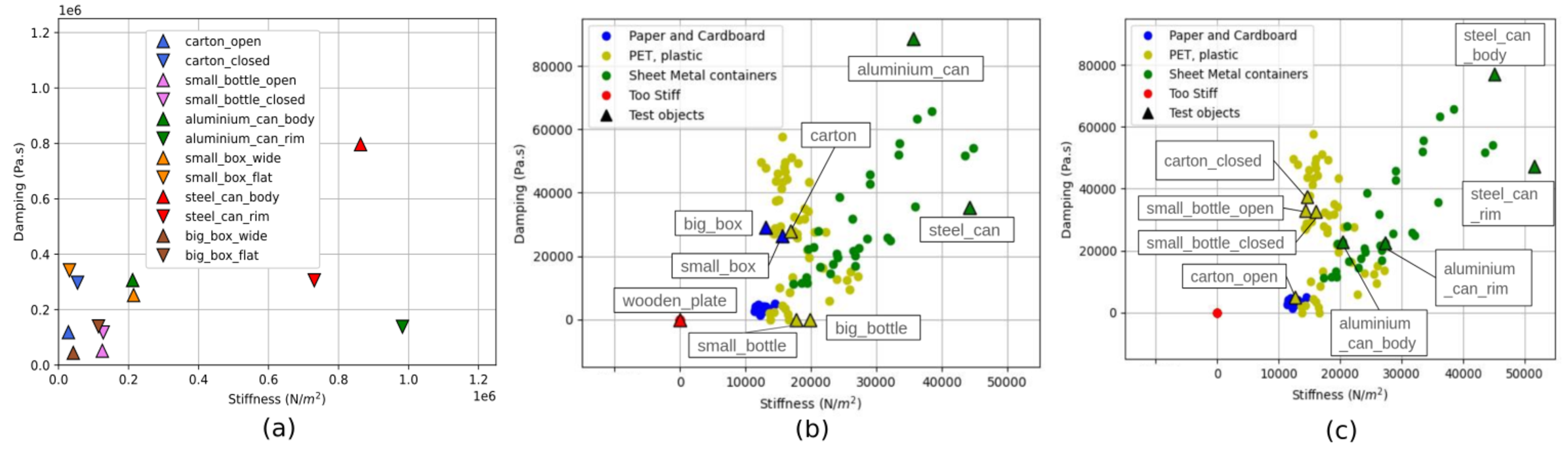}
    \caption{2-D stiffness/damping classification with all the recycling objects. (a) Using the professional setup, compression at $1.6 mm/s$. (b) Using the Kinova Gen3 arm with the Robotiq 2F-85 gripper, with different objects, compression at $1.6 mm/s$. (c) Using the robot with different grasps on the same objects, compression at $1.6 mm/s$.}
    \label{fig:recycling_objects_2d_plots}
\end{figure*}

The data presented in Table~\ref{tab:recycling_set_values} shows how the properties estimated by the Hunt-Crossley model in Eq.~\ref{eqn:HCModel} varied for irregular objects of different materials. Notably, objects made from recyclable plastics seem to have  very low $\eta$ and very high exponent $n$ based on the shape of their stress-strain curves. Further, our material categories could be separated based on ascending order of the stiffness value, which allowed us to make a practical demonstration of the usefulness of stiffness estimation in robotic systems.
The sample data from the test objects are shown in Fig.~\ref{fig:recycling_set_raw}. The unprocessed data from the Robotiq 2F-85 gripper was converted from current-position to stress-strain via Eqn.~\ref{eqn:2F85_calibration}, and with knowledge of the gripper fingers' surface area and the object's thickness. The objects' estimated properties are presented in Table~\ref{tab:recycling_set_values}, and the model fits are displayed graphically in Fig.~\ref{fig:estimation_recreation}. Fig.~\ref{fig:recycling_objects_2d_plots}(a) shows the data collected from the professional setup and the estimated stiffness and damping values of different objects. Fig.~\ref{fig:recycling_objects_2d_plots}(b) shows different test objects being categorized via stiffness estimation. Fig.~\ref{fig:recycling_objects_2d_plots}(c) shows the effects of different grasp positions on the classification task---even though the estimated values are different when objects are grasped at different locations, the values are still within range of their material categories. From Fig.~\ref{fig:recycling_objects_2d_plots}, we also observed that although the absolute values estimated by the robot are erroneous, the relative ordering of the stiffness of different materials remains consistent across both setups.

\section{Conclusion}
\label{sec:conclusion}

We have systematically experimentally evaluated the capacity of standard robot grippers to gauge material elasticity and object stiffness and the possibility of discriminating the sample objects based on their material composition. A professional biaxial compression device served as a reference. We have adapted and tested several methods to process the mechanical response curves obtained by squeezing the objects. Based on our results, we provide specific practical recommendations. The data and code are available at \url{https://osf.io/zetg3}.

\textbf{Gauging elasticity/stiffness.} We found: (i) the robot grippers, even after calibration of the force feedback, were found to have a limited capability of delivering accurate estimates of absolute values of the modulus of elasticity (Young’s modulus) at different compression levels; (ii) relative ordering of the objects based on their stiffness is largely consistent across the devices; (iii) precycling---measuring elasticity not on the first compression cycle but on the fourth or fifth as recommended by standards \cite{iso3386polymeric}---did not have a significant effect and hence data from a single compression cycle can be directly used; (iv) slower compression speeds are advantageous for estimation. Although the mechanical response stress/strain curves of the soft objects are nonlinear, in practice it proved more accurate not to attempt fitting gradients and estimating Young's modulus at different compression levels. Instead, fitting a single straight line over the whole stress/strain curve gave the most stable results.

\textbf{Gauging viscoelasticity.} Deformable objects typically have viscoelastic properties characterized by: (i) a hysteresis curve corresponding to energy dissipation during a compression and decompression cycle, and (ii) a damping term that adds the compression velocity to the stress/stain relation (see Sec.~\ref{subsec:DynMod_Hyst}). We found: (i) the area under the hysteresis curve is hard to use in practice as it requires continuous  compression and decompression, which is possible only for specific objects like foams, and accurate measuring apparatus; (ii) the damping term in the Hunt-Crossley model can be fitted online from a single compression and estimation from gripper data (Robotiq 2F-85) was sufficiently similar to the estimation from the reference setup in terms of the relative ordering of the samples' damping term. The viscoelasticity parameter can thus also be estimated from a single grasp and complement the elasticity estimation, allowing better material discrimination thanks to the second dimension.

\textbf{Online waste sorting.} We developed a mock single-stream recycling scenario that sorted objects regularly found in dry waste into their material classes by estimating their material properties.  We collected data on four different material classes \textit{Paper and Cardboard}, \textit{PET and Plastic}, \textit{Sheet Metal Containers} and \textit{Too Stiff} on with the Robotiq 2F-85 gripper. By using our best performing model, the Hunt-Crossley model, we estimated stiffness and damping for all objects and material classes, and used these estimates to successfully categorise all objects in real time.

\textbf{Practical recommendations for object discrimination based on material elasticity.} From the experiments and analyses performed we derive the following guidelines: (i) a single grasp (compression cycle) with a 2-finger robot gripper gives rise to stress/strain curves that separate even highly similar samples by their elasticity; (ii) two values (elasticity and viscoelasticity) can be estimated online from a single compression cycle facilitating object discrimination using even simple linear classifiers; (iii) robust estimates of elasticity and viscoelasticity were relative---preserving the same ordering of objects as estimated from the professional setup; absolute values of elasticity and viscoelasticity differ dramatically between setups; (iv) elasticity and viscoelasticity values of objects can be estimated online and simultaneously from a single compression cycle using the Hunt-Crossley model and allow for discriminating grocery item by their material composition; (v) slower compression speeds yielded better estimates and discrimination; (vi) precycling or compression followed by decompression did not improve estimates, suggesting that a single grasp suffices and objects can be transferred to a corresponding container immediately after or even while they are picked.


\section{Discussion and Future Work}
\label{sec:discussion}


Overall, this work provided a systematic investigation of the feasibility of online haptic discrimination of objects based on their viscoelastic properties using standard robot grippers.
All the processing we used---detecting contact, obtaining the stress/strain curve, estimating elasticity, and fitting a Hunt-Crossley model---is white-box, simple (compared to finite element method), and can work online. 
For discrimination, we used simple linear classifiers. Employing state-of-the-art classifiers will certainly increase performance (see \cite{pliska2024singlegrasp} where SVMs and LSTMs were used on partly the same data).
Visual-based methods can complement haptic data. The viscoelastic properties perceived can be further used to devise appropriate manipulation policies \cite{arriola2020modeling,gemici2014learning}. 

In terms of the robot devices we tested, the Robotiq 2F-85 gripper is limited in its sampling rate, but it has the important capability to be able continuously measure the effort during a compression and release cycle. The effort is in motor current (A) and calibration is necessary to obtain force (N) to estimate the elastic modulus in real physical units. However, even after calibration, such estimation was not accurate. On the other hand, if only relative ordering or discrimination of viscoelastic objects is desired, this device will suffice. The OnRobot RG6 in our settings could only measure force when a set threshold was exceeded. The threshold then had to be reset to continue compression. Such measurement is slow but sufficient to coarsely estimate material elasticity. However, viscoleasticity or damping cannot be estimated as the rate of compression (or decompression) is not available. For comparison, we also show data from pressing an an object with the robot flange and measuring the resistance with a force sensor at the manipulator's wrist. This proved accurate but may not be applicable in many situations when grasping an object is needed simultaneously with elasticity measurements.

\section{Declarations}

\backmatter

\bmhead{Funding}
This work was co-funded by the European Union under the project ROBOPROX (reg. no. CZ.02.01.01/00/22\_008/0004590). Hynek Chlup was supported by the Czech Science Foundation (GA CR) (no. GA23-06920S).

\bmhead{Competing Interests}
Not applicable.

\bmhead{Availibility of Data and Material}
All raw data is made available in our public \href{https://osf.io/dxsg8/?view_only=979775a79d934a0083a1b2008544183e}{data repository}.

\bmhead{Code Availibility}
Code for the analysis of data is made available in our public \href{https://osf.io/69xjw/?view_only=979775a79d934a0083a1b2008544183e}{code repository}.

\bmhead{Ethics Approval}
Not applicable.
\bmhead{Consent to Participate}
Not applicable.
\bmhead{Consent for Publication}
Not applicable.

\bmhead{Author Contributions}
All authors contributed to the study conception and design. Material preparation, data collection and analysis were performed by Shubhan P. Patni and Pavel Stoudek. The first draft of the manuscript was written by Shubhan P. Patni and Pavel Stoudek and all authors commented on previous versions of the manuscript. All authors read and approved the final manuscript.

\bmhead{Acknowledgements}
We would like to thank Karla Stepanova for her inputs, Bedrich Himmel for assistance with gripper calibration, and Jiri Hartvich for early code for online material classification. 





\bibliography{sn-bibliography}

\begin{thebibliography}{10}
\expandafter\ifx\csname url\endcsname\relax
  \def\url#1{\burl{#1}}\fi
\expandafter\ifx\csname urlprefix\endcsname\relax\def\urlprefix{URL }\fi
\providecommand{\bibinfo}[2]{#2}
\providecommand{\eprint}[2][]{\url{#2}}
\providecommand{\doi}[1]{\url{https://doi.org/#1}}
\bibcommenthead

\bibitem{sanchez2018robotic}
\bibinfo{author}{Sanchez, J.}, \bibinfo{author}{Corrales, J.-A.},
  \bibinfo{author}{Bouzgarrou, B.-C.} \& \bibinfo{author}{Mezouar, Y.}
\newblock \bibinfo{title}{Robotic manipulation and sensing of deformable
  objects in domestic and industrial applications: a survey}.
\newblock \emph{\bibinfo{journal}{The International Journal of Robotics
  Research}} \textbf{\bibinfo{volume}{37}}, \bibinfo{pages}{688--716}
  (\bibinfo{year}{2018}).

\bibitem{lin2023non}
\bibinfo{author}{Lin, J.} \emph{et~al.}
\newblock \bibinfo{title}{Non-destructive fruit firmness evaluation using a
  soft gripper and vision-based tactile sensing}.
\newblock \emph{\bibinfo{journal}{Computers and Electronics in Agriculture}}
  \textbf{\bibinfo{volume}{214}}, \bibinfo{pages}{108256}
  (\bibinfo{year}{2023}).

\bibitem{ribeiro2020fruit}
\bibinfo{author}{Ribeiro, P.}, \bibinfo{author}{Cardoso, S.},
  \bibinfo{author}{Bernardino, A.} \& \bibinfo{author}{Jamone, L.}
\newblock \bibinfo{editor}{IEEE} (ed.) \emph{\bibinfo{title}{Fruit quality
  control by surface analysis using a bio-inspired soft tactile sensor}}.
\newblock (ed.\bibinfo{editor}{IEEE}) \emph{\bibinfo{booktitle}{2020 IEEE/RSJ
  International Conference on Intelligent Robots and Systems (IROS)}},
  \bibinfo{pages}{8875--8881} (\bibinfo{organization}{IEEE},
  \bibinfo{year}{2020}).

\bibitem{cardin2023gripper}
\bibinfo{author}{Cardin-Catalan, D.}, \bibinfo{author}{Morales, A.},
  \bibinfo{author}{Llop-Harillo, I.}, \bibinfo{author}{Perez-Gonzalez, A.} \&
  \bibinfo{author}{del Pobil, A.~P.}
\newblock \bibinfo{title}{A gripper for delicate edible manipulation}.
\newblock \emph{\bibinfo{journal}{Engineering Science and Technology, an
  International Journal}} \textbf{\bibinfo{volume}{47}},
  \bibinfo{pages}{101537} (\bibinfo{year}{2023}).

\bibitem{lubongo2022assessment}
\bibinfo{author}{Lubongo, C.} \& \bibinfo{author}{Alexandridis, P.}
\newblock \bibinfo{title}{Assessment of performance and challenges in use of
  commercial automated sorting technology for plastic waste}.
\newblock \emph{\bibinfo{journal}{Recycling}} \textbf{\bibinfo{volume}{7}},
  \bibinfo{pages}{11} (\bibinfo{year}{2022}).

\bibitem{chin2019automated}
\bibinfo{author}{Chin, L.}, \bibinfo{author}{Lipton, J.},
  \bibinfo{author}{Yuen, M.~C.}, \bibinfo{author}{Kramer-Bottiglio, R.} \&
  \bibinfo{author}{Rus, D.}
\newblock \bibinfo{editor}{Hosoda, K.} (ed.) \emph{\bibinfo{title}{Automated
  recycling separation enabled by soft robotic material classification}}.
\newblock (ed.\bibinfo{editor}{Hosoda, K.}) \emph{\bibinfo{booktitle}{2019 2nd
  IEEE International Conference on Soft Robotics (RoboSoft)}},
  Vol.~\bibinfo{volume}{2}, \bibinfo{pages}{102--107}
  (\bibinfo{organization}{IEEE}, \bibinfo{year}{2019}).

\bibitem{papadopoulos2023deformable}
\bibinfo{author}{Papadopoulos, G.} \emph{et~al.}
\newblock \bibinfo{title}{On deformable object handling: multi-tool
  end-effector for robotized manipulation and layup of fabrics and composites}.
\newblock \emph{\bibinfo{journal}{The International Journal of Advanced
  Manufacturing Technology}} \textbf{\bibinfo{volume}{128}},
  \bibinfo{pages}{1675--1687} (\bibinfo{year}{2023}).

\bibitem{zhang2023determination}
\bibinfo{author}{Zhang, S.} \& \bibinfo{author}{Zhang, Y.}
\newblock \bibinfo{title}{Determination method of stable grasping parameters
  for irregular sheet sorting}.
\newblock \emph{\bibinfo{journal}{The International Journal of Advanced
  Manufacturing Technology}} \textbf{\bibinfo{volume}{128}},
  \bibinfo{pages}{2075--2085} (\bibinfo{year}{2023}).

\bibitem{liu2018variable}
\bibinfo{author}{Liu, L.}, \bibinfo{author}{Zhang, Y.}, \bibinfo{author}{Liu,
  G.} \& \bibinfo{author}{Xu, W.}
\newblock \bibinfo{title}{Variable motion mapping to enhance stiffness
  discrimination and identification in robot hand teleoperation}.
\newblock \emph{\bibinfo{journal}{Robotics and Computer-Integrated
  Manufacturing}} \textbf{\bibinfo{volume}{51}}, \bibinfo{pages}{202--208}
  (\bibinfo{year}{2018}).

\bibitem{scimeca2022action}
\bibinfo{author}{Scimeca, L.} \emph{et~al.}
\newblock \bibinfo{title}{Action augmentation of tactile perception for
  soft-body palpation}.
\newblock \emph{\bibinfo{journal}{Soft robotics}} \textbf{\bibinfo{volume}{9}},
  \bibinfo{pages}{280--292} (\bibinfo{year}{2022}).

\bibitem{osf}
\bibinfo{author}{Patni, S.}, \bibinfo{author}{Stoudek, P.} \&
  \bibinfo{author}{Hoffmann, M.}
\newblock \bibinfo{title}{Squeezing data and processing for mixed set}
  (\bibinfo{year}{2021}).
\newblock
  \urlprefix\url{https://osf.io/gec6s/?view_only=979775a79d934a0083a1b2008544183e}.

\bibitem{dataset}
\bibinfo{author}{Stoudek, P.}
\newblock \bibinfo{title}{Raw data from squeezing on deformable objects}
  (\bibinfo{year}{2020}).
\newblock
  \urlprefix\url{https://drive.google.com/drive/folders/16sUG-zCwCg5HIF7EbCTBqgsklw7-vOpg?usp=sharing}.

\bibitem{li2020review}
\bibinfo{author}{Li, Q.} \emph{et~al.}
\newblock \bibinfo{title}{A review of tactile information: Perception and
  action through touch}.
\newblock \emph{\bibinfo{journal}{IEEE Transactions on Robotics}}
  (\bibinfo{year}{2020}).

\bibitem{luo2017robotic}
\bibinfo{author}{Luo, S.}, \bibinfo{author}{Bimbo, J.},
  \bibinfo{author}{Dahiya, R.} \& \bibinfo{author}{Liu, H.}
\newblock \bibinfo{title}{Robotic tactile perception of object properties: A
  review}.
\newblock \emph{\bibinfo{journal}{Mechatronics}} \textbf{\bibinfo{volume}{48}},
  \bibinfo{pages}{54--67} (\bibinfo{year}{2017}).

\bibitem{spiers2016single}
\bibinfo{author}{Spiers, A.~J.}, \bibinfo{author}{Liarokapis, M.~V.},
  \bibinfo{author}{Calli, B.} \& \bibinfo{author}{Dollar, A.~M.}
\newblock \bibinfo{title}{Single-grasp object classification and feature
  extraction with simple robot hands and tactile sensors}.
\newblock \emph{\bibinfo{journal}{IEEE transactions on haptics}}
  \textbf{\bibinfo{volume}{9}}, \bibinfo{pages}{207--220}
  (\bibinfo{year}{2016}).

\bibitem{delgado2015tactile}
\bibinfo{author}{Delgado, A.}, \bibinfo{author}{Jara, C.~A.},
  \bibinfo{author}{Mira, D.} \& \bibinfo{author}{Torres, F.}
\newblock \bibinfo{editor}{Filipe, J.} (ed.) \emph{\bibinfo{title}{A
  tactile-based grasping strategy for deformable objects' manipulation and
  deformability estimation}}.
\newblock (ed.\bibinfo{editor}{Filipe, J.}) \emph{\bibinfo{booktitle}{12th
  International Conference on Informatics in Control, Automation and Robotics
  (ICINCO)}}, \bibinfo{pages}{369--374} (\bibinfo{organization}{IEEE},
  \bibinfo{year}{2015}).

\bibitem{hosoda2010robust}
\bibinfo{author}{Hosoda, K.} \& \bibinfo{author}{Iwase, T.}
\newblock \bibinfo{editor}{Luo, R.} (ed.) \emph{\bibinfo{title}{Robust haptic
  recognition by anthropomorphic bionic hand through dynamic interaction}}.
\newblock (ed.\bibinfo{editor}{Luo, R.}) \emph{\bibinfo{booktitle}{2010
  IEEE/RSJ International Conference on Intelligent Robots and Systems}},
  \bibinfo{pages}{1236--1241} (\bibinfo{organization}{IEEE},
  \bibinfo{year}{2010}).

\bibitem{gemici2014learning}
\bibinfo{author}{Gemici, M.~C.} \& \bibinfo{author}{Saxena, A.}
\newblock \bibinfo{editor}{Lynch, K.} (ed.) \emph{\bibinfo{title}{Learning
  haptic representation for manipulating deformable food objects}}.
\newblock (ed.\bibinfo{editor}{Lynch, K.}) \emph{\bibinfo{booktitle}{2014
  IEEE/RSJ International Conference on Intelligent Robots and Systems}},
  \bibinfo{pages}{638--645} (\bibinfo{organization}{IEEE},
  \bibinfo{year}{2014}).

\bibitem{scimeca2019non}
\bibinfo{author}{Scimeca, L.} \emph{et~al.}
\newblock \bibinfo{editor}{Dudek, G.} (ed.)
  \emph{\bibinfo{title}{Non-destructive robotic assessment of mango ripeness
  via multi-point soft haptics}}.
\newblock (ed.\bibinfo{editor}{Dudek, G.}) \emph{\bibinfo{booktitle}{2019
  International Conference on Robotics and Automation (ICRA)}},
  \bibinfo{pages}{1821--1826} (\bibinfo{organization}{IEEE},
  \bibinfo{year}{2019}).

\bibitem{wang2021tactual}
\bibinfo{author}{Wang, L.}, \bibinfo{author}{Li, Q.}, \bibinfo{author}{Lam, J.}
  \& \bibinfo{author}{Wang, Z.}
\newblock \bibinfo{title}{Tactual recognition of soft objects from deformation
  cues}.
\newblock \emph{\bibinfo{journal}{IEEE Robotics and Automation Letters}}
  \textbf{\bibinfo{volume}{7}}, \bibinfo{pages}{96--103}
  (\bibinfo{year}{2022}).

\bibitem{smardzewski2008nonlinear}
\bibinfo{author}{Smardzewski, J.}, \bibinfo{author}{Grbac, I.} \&
  \bibinfo{author}{Prekrat, S.}
\newblock \bibinfo{title}{Nonlinear mechanics of hyper elastic polyurethane
  furniture foams}.
\newblock \emph{\bibinfo{journal}{Drvna industrija: Znanstveni {\v{c}}asopis za
  pitanja drvne tehnologije}} \textbf{\bibinfo{volume}{59}},
  \bibinfo{pages}{23--28} (\bibinfo{year}{2008}).

\bibitem{iso3386polymeric}
\bibinfo{title}{{Polymeric materials, cellular flexible -- Determination of
  stress-strain characteristics in compression, Part 1: Low-density materials}}
  (\bibinfo{year}{1986}).
\newblock
  \urlprefix\url{https://www.iso.org/obp/ui/#iso:std:iso:3386:-1:ed-2:v1:en}.

\bibitem{frank2010learning}
\bibinfo{author}{Frank, B.}, \bibinfo{author}{Schmedding, R.},
  \bibinfo{author}{Stachniss, C.}, \bibinfo{author}{Teschner, M.} \&
  \bibinfo{author}{Burgard, W.}
\newblock \bibinfo{editor}{Luo, R.} (ed.) \emph{\bibinfo{title}{Learning the
  elasticity parameters of deformable objects with a manipulation robot}}.
\newblock (ed.\bibinfo{editor}{Luo, R.}) \emph{\bibinfo{booktitle}{2010
  IEEE/RSJ International Conference on Intelligent Robots and Systems}},
  \bibinfo{pages}{1877--1883} (\bibinfo{organization}{IEEE},
  \bibinfo{year}{2010}).

\bibitem{longhini2022edo}
\bibinfo{author}{Longhini, A.} \emph{et~al.}
\newblock \bibinfo{editor}{Althoefer, K.} (ed.) \emph{\bibinfo{title}{Edo-net:
  Learning elastic properties of deformable objects from graph dynamics}}.
\newblock (ed.\bibinfo{editor}{Althoefer, K.}) \emph{\bibinfo{booktitle}{2023
  IEEE International Conference on Robotics and Automation (ICRA 2023)}},
  \bibinfo{pages}{3875--3881} (\bibinfo{year}{2023}).

\bibitem{longhini2023elastic}
\bibinfo{author}{Longhini, A.} \emph{et~al.}
\newblock \bibinfo{editor}{Althoefer, K.} (ed.) \emph{\bibinfo{title}{Elastic
  context: Encoding elasticity for data-driven models of textiles}}.
\newblock (ed.\bibinfo{editor}{Althoefer, K.}) \emph{\bibinfo{booktitle}{2023
  IEEE International Conference on Robotics and Automation (ICRA 2023)}},
  \bibinfo{pages}{1764--1770} (\bibinfo{year}{2023}).

\bibitem{narang2021sim}
\bibinfo{author}{Narang, Y.}, \bibinfo{author}{Sundaralingam, B.},
  \bibinfo{author}{Macklin, M.}, \bibinfo{author}{Mousavian, A.} \&
  \bibinfo{author}{Fox, D.}
\newblock \bibinfo{title}{Sim-to-real for robotic tactile sensing via
  physics-based simulation and learned latent projections}.
\newblock \emph{\bibinfo{journal}{arXiv preprint arXiv:2103.16747}}
  (\bibinfo{year}{2021}).

\bibitem{bickel2009capture}
\bibinfo{author}{Bickel, B.} \emph{et~al.}
\newblock \bibinfo{title}{Capture and modeling of non-linear heterogeneous soft
  tissue}.
\newblock \emph{\bibinfo{journal}{ACM Transactions on Graphics (TOG)}}
  \textbf{\bibinfo{volume}{28}}, \bibinfo{pages}{1--9} (\bibinfo{year}{2009}).

\bibitem{zaidi2017model}
\bibinfo{author}{Zaidi, L.}, \bibinfo{author}{Corrales, J.~A.},
  \bibinfo{author}{Bouzgarrou, B.~C.}, \bibinfo{author}{Mezouar, Y.} \&
  \bibinfo{author}{Sabourin, L.}
\newblock \bibinfo{title}{Model-based strategy for grasping 3d deformable
  objects using a multi-fingered robotic hand}.
\newblock \emph{\bibinfo{journal}{Robotics and Autonomous Systems}}
  \textbf{\bibinfo{volume}{95}}, \bibinfo{pages}{196--206}
  (\bibinfo{year}{2017}).

\bibitem{haddadi2012real}
\bibinfo{author}{Haddadi, A.} \& \bibinfo{author}{Hashtrudi-Zaad, K.}
\newblock \bibinfo{title}{Real-time identification of hunt--crossley dynamic
  models of contact environments}.
\newblock \emph{\bibinfo{journal}{IEEE Transactions on Robotics}}
  \textbf{\bibinfo{volume}{28}}, \bibinfo{pages}{555--566}
  (\bibinfo{year}{2012}).

\bibitem{bednarek2019robotic}
\bibinfo{author}{Bednarek, J.}, \bibinfo{author}{Bednarek, M.},
  \bibinfo{author}{Kicki, P.} \& \bibinfo{author}{Walas, K.}
\newblock \bibinfo{editor}{Hosoda, K.} (ed.) \emph{\bibinfo{title}{Robotic
  touch: Classification of materials for manipulation and walking}}.
\newblock (ed.\bibinfo{editor}{Hosoda, K.}) \emph{\bibinfo{booktitle}{2019 2nd
  IEEE International Conference on Soft Robotics (RoboSoft)}},
  \bibinfo{pages}{527--533} (\bibinfo{organization}{IEEE},
  \bibinfo{year}{2019}).

\bibitem{yao2023estimating}
\bibinfo{author}{Yao, S.} \& \bibinfo{author}{Hauser, K.}
\newblock \bibinfo{editor}{Althoefer, K.} (ed.)
  \emph{\bibinfo{title}{Estimating tactile models of heterogeneous deformable
  objects in real time}}.
\newblock (ed.\bibinfo{editor}{Althoefer, K.}) \emph{\bibinfo{booktitle}{2023
  IEEE International Conference on Robotics and Automation (ICRA 2023)}},
  \bibinfo{pages}{12583--12589} (\bibinfo{year}{2023}).

\bibitem{giancoli1995physics}
\bibinfo{author}{Giancoli, D.~C.}
\newblock \emph{\bibinfo{title}{Physics}} \bibinfo{edition}{Fourth} edn
  (\bibinfo{publisher}{Prentice-Hall International}, \bibinfo{year}{1995}).

\bibitem{Stoudek2020}
\bibinfo{author}{Stoudek, P.}
\newblock \emph{\bibinfo{title}{Extracting material properties of objects from
  haptic exploration using multiple robotic grippers}}.
\newblock Master's thesis, \bibinfo{school}{Czech Technical University, Faculty
  of Electrical Engineering} (\bibinfo{year}{2020}).

\bibitem{gitlab-repo-ipalm}
\bibinfo{author}{Stoudek, P.} \& \bibinfo{author}{Mareš, M.}
  (\bibinfo{year}{2021}).
\newblock
  \urlprefix\url{https://gitlab.fel.cvut.cz/body-schema/ipalm/ipalm-grasping}.

\bibitem{pliska2024singlegrasp}
\bibinfo{author}{Pliska, M.} \emph{et~al.}
\newblock \bibinfo{title}{Single-grasp deformable object discrimination: the
  effect of gripper morphology, sensing modalities, and action parameters}
  (\bibinfo{year}{2024}).
\newblock \eprint{2204.06343}.

\bibitem{arriola2020modeling}
\bibinfo{author}{Arriola-Rios, V.~E.} \emph{et~al.}
\newblock \bibinfo{title}{Modeling of deformable objects for robotic
  manipulation: A tutorial and review}.
\newblock \emph{\bibinfo{journal}{Frontiers in Robotics and AI}}
  \textbf{\bibinfo{volume}{7}}, \bibinfo{pages}{82} (\bibinfo{year}{2020}).

\end{thebibliography}

\end{document}